\documentclass{article}

\usepackage{PRIMEarxiv}

\usepackage[utf8]{inputenc} 
\usepackage[T1]{fontenc}    
\usepackage{hyperref}       
\usepackage{url}            
\usepackage{booktabs}       
\usepackage{amsfonts}       
\usepackage{nicefrac}       
\usepackage{microtype}      
\usepackage{lipsum}
\usepackage{fancyhdr}       
\usepackage{graphicx}       
\graphicspath{{media/}}     

\usepackage{amsmath,amsfonts,amssymb,amsthm,dsfont,stmaryrd}
\usepackage{bbold}
\usepackage{booktabs}       
\usepackage{microtype}      
\usepackage[dvipsnames]{xcolor}
\hypersetup{
	colorlinks,
	linkcolor={red!50!black},
	citecolor={blue!50!black},
	urlcolor={blue!80!black}
}

\usepackage{url}
\usepackage{cite}
\usepackage{graphicx}
\usepackage[mathlines,switch]{lineno}

\usepackage{caption}
\usepackage{subcaption}

\usepackage[linesnumbered,ruled,vlined]{algorithm2e}

\usepackage{afterpage}

\usepackage{wrapfig}
\usepackage{multirow}

\newtheorem{theorem}{Theorem}
\newtheorem{remark}{Remark}

\newtheorem{example}{Example}
\newtheorem{definition}{Definition}

\usepackage[capitalize,nosort]{cleveref}
\crefname{section}{Sec.}{Secs.}
\Crefname{section}{Section}{Sections}
\Crefname{table}{Table}{Tables}
\crefname{table}{Tab.}{Tabs.}
\Crefname{appsec}{Appendix}{Appendices}
\crefname{appsec}{App.}{Apps.}

\def\e{\varepsilon}

\def\R{\mathbb{R}}

\def\e{\varepsilon}

\def\G{\mathcal{G}}

\def\ri{\mathcal{Q}}

\def\1{\mathbb{1}}

\def\nor{\mathcal{N}}

\newcommand\il[1]{\langle #1 \rangle}

\def\cost{\mathrm{cost}}

\def\R{\mathbb{R}}

\title{Feature-Based Interpolation and Geodesics in the \\ Latent Spaces of Generative Models
}

\author{
  \L{}ukasz Struski,\quad Micha\l{} Sadowski,\quad Tomasz Danel,\quad Jacek Tabor,\quad Igor T. Podolak \\[1em]
  The Faculty of Mathematics and Computer Science,\\
   Jagiellonian University, Krak\'ow, Poland \\
}

\begin{document}
\maketitle

\begin{abstract}
	Interpolating between points is a problem connected simultaneously with finding geodesics and study of generative models. In the case of geodesics, we search for the curves with the shortest length, while in the case of generative models we typically apply linear interpolation in the latent space.
	However, this interpolation uses implicitly the fact that Gaussian is unimodal. Thus the problem of interpolating in the case when the latent density is non-Gaussian is an open problem.
	
	In this paper, we present a general and unified approach to interpolation, which simultaneously allows us to search for geodesics and interpolating curves in latent space in the case of arbitrary density.
	Our results have a strong theoretical background based on the introduced quality measure of an interpolating curve. In particular, we show that maximising the quality measure of the curve can be equivalently understood as a search of geodesic for a certain redefinition of the Riemannian metric on the space. 
	
	We provide examples in three important cases. First, we show that our approach can be easily applied to finding geodesics on manifolds. Next, 
	we focus our attention in finding interpolations in pre-trained generative models. We show that our model effectively works in the case of arbitrary density. Moreover, we can interpolate in the subset of the space consisting of data possessing a given feature. The last case is focused on finding interpolation in the space of chemical compounds.
\end{abstract}

\section{Introduction}

Generative models now play both theoretical and practical roles in machine learning~\cite{jabbar2020survey,walters2020assessing,li2021}. Besides creating pleasing and realistic images, they can be used to generate new objects that match the probability distribution of the input data manifold, leading to the discovery of new knowledge and relations in the data. There are numerous applications of generative models, e.g. missing data imputation~\cite{yu2018generative,yoon2018gain}, text translation~\cite{vaswani2017attention,devlin2018bert}, and discovery of new molecules~\cite{Gomez-Bombarelli2017,ingraham2021generative,sanchez2018inverse}. They often come with latent space representations that can help, e.g., with speech generation and comprehension~\cite{oord2016wavenet,raymond2007} and EEG signal transformations~\cite{cheng2020subjectaware,hartmann2018eeggan} and can be used as a tool to discover meaningful paths between represented objects. Recently, deep generative models were also proposed to solve continual learning problems~\cite{shin2017,zhai2019lifelong} and to increase the safety of autonomous driving~\cite{zhang2018deeproad}.


Interpolations are a powerful tool that is employed to assess the quality and robustness of generative models. During training, many such models create a latent representation of the input data. For example, auto-encoders~\cite{KingmaWelling2013,tolstikhin2017wasserstein} encode the data in a lower-dimensional latent space, and generative adversarial networks (GANs)~\cite{Goodfellow2014} use vectors sampled from a latent space to generate instances that mimic the input data. By choosing data points on the interpolation curves between two selected points in the latent space, one can evaluate several important properties of a generative system. For instance, these intermediate points should correspond to realistic objects that are highly similar to the training data if the model generalises well. Furthermore, in well-represented latent spaces, the transition from one object to another is gradual, e.g. the changes of an object depicted in images decoded along the interpolation path are proportional to their distance in the latent space. The ability to manipulate the interpolating function in the latent spaces preserving the above-mentioned characteristics enables efficient exploration of the input data space. This in turn leads to more meaningful, novel, and optimal solutions for creative tasks, e.g. molecular design\cite{Gomez-Bombarelli2017} or path-finding in robotics~\cite{detlefsen2019explicit,radford2015unsupervised, White2016sampling, LesniakSieradzkiPodolak:2019}. 
Our proposition can be used in extended tasks, where the problem is to reconstruct an unknown data from a number of samples, i.e. compressive sensing~\cite{bora2017compressedsensing, asim2020invertible}. The generative models give a very structured input space distribution representation, much superior to simple neighbourhood sampling. Samples from a path found with our algorithm shall have all the important data characteristics~\cite{adler2018deepbayesian}.

\begin{figure*}[t]
	\centering
	\includegraphics[width=0.75\textwidth]{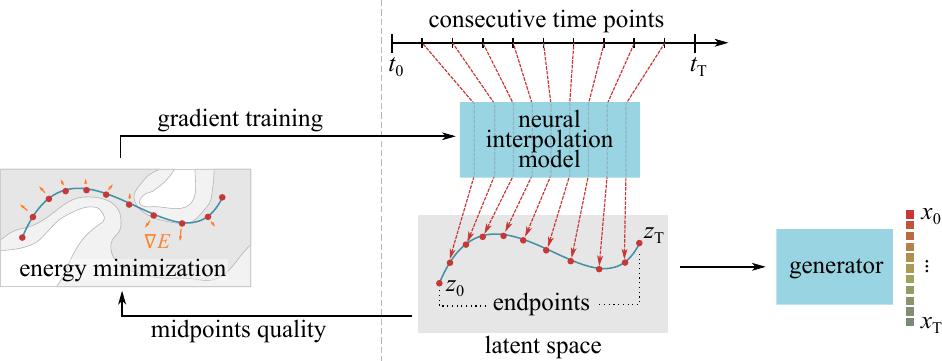}
	\caption{
		Our interpolation network maps a sequence of points, one at a time, from the $(t_0,\dots,t_T)$ interval, where the endpoints $t_0$ and $t_T$ correspond to the queried input data objects between which the interpolation is performed. The interpolation network minimises the energy function of the whole induced curve, based on a pre-defined quality function, e.g. data probability density.
	}
	\label{fig:our_model}
\end{figure*}


There are a number of approaches to constructing interpolations in latent spaces. A minimal path approach may use a local Riemannian structure metric~\cite{Agustsson2018, arvanitidis2018} or an embedding~\cite{LesniakSieradzkiPodolak:2019}. Some propose to linearly interpolate using a $w$ vector computed as a mean latent space difference for objects with and without some feature~\cite{upchurch2017deep}. Other methods require that all the objects on the constructed path are interpretable in the model context, i.e. preserve the characteristics of the input data, or fulfil other requirements, e.g. in a chemical model all molecules should be synthesizable~\cite{gao2020synthesizability,prykhodko2019adenovo}. In order to generate realistic samples, it should be ensured that the generated path crosses high probability regions in the data space, according to the manifold hypothesis~\cite{Bengio:2013,Fefferman:2016,connor:2021pmlr}. A related class of methods is latent space disentanglement and attribute factorisation, but its objective is different than path selection~\cite{burgess2018understanding,lee2018diverseimagetoimageECCV,kim2018disentangling, detlefsen2019explicit}.
One must remember that the latent space cannot be simply treated as Euclidean, and thus linear interpolations may deviate from the data distribution manifold mapped, in particular in problems that are not strictly continuous, as e.g. cheminformatics~\cite{michelis2021on,LesniakSieradzkiPodolak:2019,arvanitidis2018,Gomez-Bombarelli2017}. Hence our proposition to base the induced latent metric on the feature probability density of the data.

\begin{figure}[hbt]
	\centering
	\includegraphics[width=0.37\textwidth]{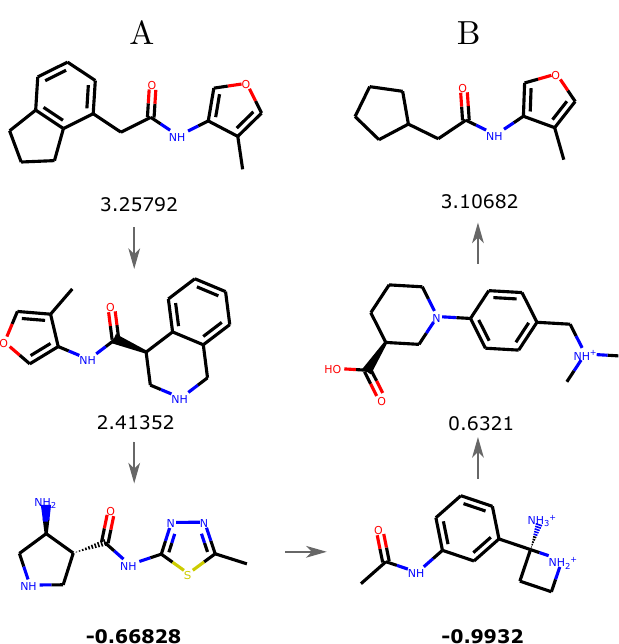}
	\caption{Exemplary optimised path between compound A and compound B that minimises octanol-water partition coefficient log\,P (values shown below the compounds). Our method optimises the interpolation path in the latent space of a trained chemical generative model to find similar compounds with the minimised log\,P value. More details are described in \cref{sec:chemical_design}.}
	\label{fig:mols_u}
\end{figure}

\begin{figure*}[htb!]
	\centering
	\renewcommand{\arraystretch}{0.1}
	\begin{tabular}{ @{}r@{\;}c@{\;}l@{} }
		\multirow{2}{*}{\rotatebox{90}{\makebox[10pt][c]{Pioneer}}} & \includegraphics[width=.95\textwidth]{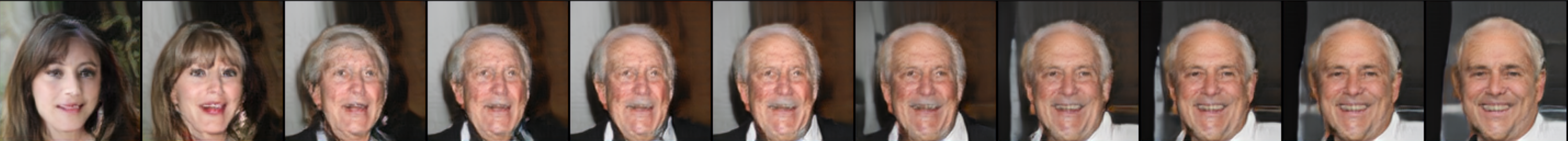}
		& \rotatebox{90}{\makebox[40pt][c]{old}} \\
		& \includegraphics[width=.95\textwidth]{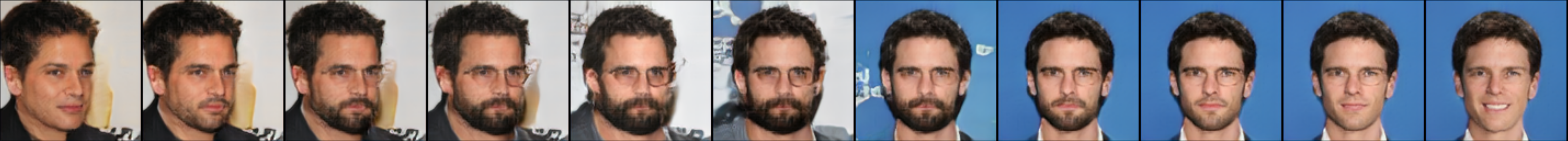}
		& \rotatebox{90}{\makebox[40pt][c]{beard}} \\
		\multirow{2}{*}{\rotatebox{90}{\makebox[10pt][c]{Glow}}}  & \includegraphics[width=.95\textwidth]{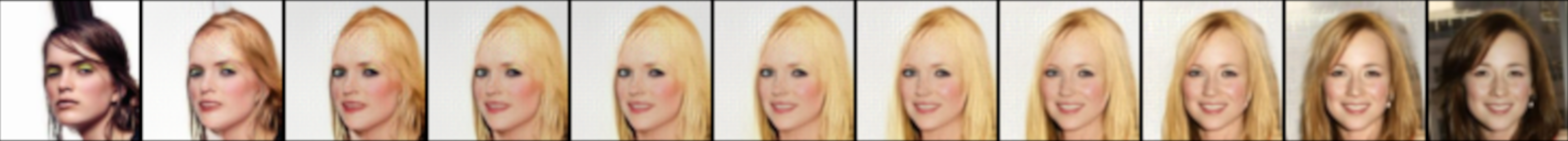}
		& \rotatebox{90}{\makebox[40pt][c]{blond hair}} \\
		& \includegraphics[width=.95\textwidth]{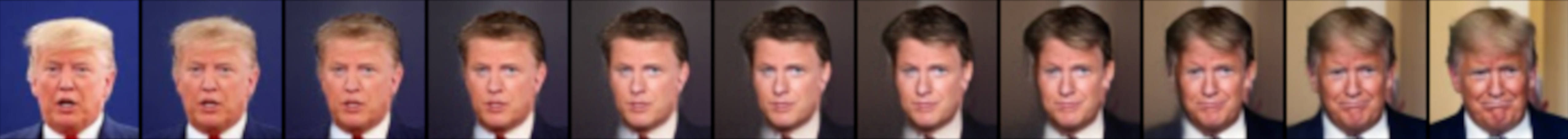}
		& \rotatebox{90}{\makebox[40pt][c]{young}} \\
	\end{tabular}
	
	\caption{Interpolations using two different generative models. A~pretrained Pioneer is used in the first two rows, and a~pretrained Glow model is used for examples in the last rows. In both cases, the endpoint images are different.  See more examples in \cref{app.add_results}}
	\label{fig:interpolation}
\end{figure*}





The objective of this paper is to present a~general approach that can be easily adapted to any generative model. It utilises kernel methodology to improve interpolation paths over simple interpolation methods, e.g. linear or spherical ones.
\cref{fig:our_model} shows the basic architecture of our approach. The scheme aims to produce short-length paths together, skipping low-density regions and fulfilling some predefined stipulations of the path elements. In particular, the methodology should produce interpolations that pass through a region representing the given feature, e.g.~see \cref{fig:interpolation,fig:pioneer_more} or optimise a feature when traversing a latent space of a system encoding chemical molecules, see \cref{fig:mols_u}. Moreover, contrary to the classical interpolations (e.g.~linear or spherical interpolations), we are able to use our approach in the case when the endpoints coincide, see \cref{fig:interpolation_the_same_image,fig:pionier_same_endpoints_more}. One of the practical solutions of the proposed model is its application to the discovery of new drugs, which requires effective methods of traversing the chemical latent space. Finding novel potent molecules involves a multi-parametric optimisation to ensure high bio-activity, lack of toxicity, and good distribution in the human body. This can be facilitated by our method that enables exploration of the latent space in the proximity of the known compounds. We show experimentally that this method works on various kinds of models and datasets. We present results on DCGAN~\cite{radford2015unsupervised}, Pioneer~\cite{heljakka2018pioneer}, StackGAN~\cite{zhang2017stackgan}, Glow models~\cite{Kingma2018glow}, and SVHN~\cite{netzer2011svhn}, MNIST~\cite{lecun-mnisthandwrittendigit-2010}, CelebA~\cite{liu2015faceattributes}, CUB-200~\cite{WelinderEtal2010}, and chemical molecule datasets~\cite{irwin2005zinc}. The approach can also be used in the case when the prior density is not Gaussian, see 
\cref{fig:semicircle_prior}.
We can also easily interpolate between measures (represented by samples), see \cref{fig:space_measure}.


To the best of our knowledge, there are no approaches that would unify latent space interpolations with geodesics and use features that depend on the current context, i.e. other features of an object, simultaneously and using kernel approach. However, there are propositions closely related to our interpolation method. The notion of the pull-back metric being induced by the class probability, is clearly stated in~\cite{michelis2021on}. A number of authors analyse the latent space generated by the variational auto-encoder VAE, e.g.~\cite{arvanitidis2018}. They show that a Riemannian metric imbues this latent space and that this metric can be evaluated in terms of mean and variance functions of the corresponding VAE by using an RBF network and show the superiority of this approach.

Others also investigate the Riemannian geometry of manifolds generated with VAE~\cite{shao2018riemannian}. Using geodesic curves for the distance between points, they develop a method for translation of a tangent vector along a path on the manifold. A change in one data point can then be transported into a semantically similar change of another. As a step forward, the proposition in this paper is aimed at finding geodesics with set data space features. 
A number of papers point out the very important characteristic, that in almost all cases the latent space cannot be treated as Euclidean and the metric is induced with $\mathbb{J}^T\mathbb{J}$, where Jacobian $\mathbb{J}=\partial g/\partial x$ and $g$ is a generator~\cite{arvanitidis2018,laine18featurebased,michelis2021on}. Therefore an $L_2$ metric used for path interpolation will, in fact, deteriorate the results. In this approach, we also postulate an energy minimisation approach, which implicitly defines the distance between the objects. The distance is now based on probability density of the data, rather than the Euclidean distance within the latent.

\textbf{Contribution} The main contribution of this paper is a framework unifying the approaches mentioned in the previous paragraph by employing kernel methods generality. We use kernels since they enable us to apply any metric in the probabilistic latent space. The intuition behind our model is to find the shortest path in a latent space of a pretrained model. The points along the path should fulfil additional restrictions specific to the problem at hand. In details, our solution is composed of the following propositions
\begin{itemize}
	\setlength{\itemsep}{1pt}
	\item a~kernel $k(\cdot,\cdot)$  defined in space $Z$ that might be given by an embedding,
	\item a~$\ri:\R^D\rightarrow[0,1]$ to assess the path elements quality using additional requirements,
	\item an interpolation model that returns a parameterised curve with parameter $t$, i.e. a time parameter that is differentiable together with a kernel changing $t$,
	\item a model that is able to work on any manifold, in particular a mixture of densities,
	\item a model that is able to produce "cycle" interpolations, i.e. with end point equal to start and passing through data and/or feature defined densities,
	\item our interpolation model that can take any pretrained generative network to work upon; thus, we can discover more meaningful representation of the data or new relations in the already existing model without changing its weights.
	
\end{itemize}
Important theoretical support of our method is given in \cref{th1}. We show that the search for optimal interpolations can be defined as searching for geodesics in a respective Riemannian space.

We also consider a case where the kernel $k$ is changing with the interpolation curve parameter $t$, so that path is constructed to be short and produce high-quality value $\ri$ objects, see e.g. \cref{fig:interpolation,fig:pioneer_more}. In the proposed approach, the selected feature is modified along the path to give results similar to a partial disentanglement of the latent space, which was not built into the original model or imposed by its training procedure. Moreover, the created path is continuous as a respectively trained neural network gives it.

We show that our kernel-based methodology gives outstanding results and can be applied to generative models already trained even when the optimisation of interpolation was not the objective of the training of this generative model.
It is presented in multiple experiments in the paper, where several architectures like DCGAN, Glow, Pioneer, and StackGAN are considered and with various datasets like MNIST, SVHN, CelebA, CUB-200, and molecular data. 

We published the source code of our method in \url{https://github.com/gmum/feature-based-interpolation}.


\section{Quality measure of a~curve in kernel space}
\label{sec:quality-measure}
We shall introduce now the latent space curve $\gamma$, which corresponds to the interpolation we are to find for a trained generative model. For this purpose, we shall use the notion of a kernel that enables us to operate in a high-dimensional feature space without ever computing the coordinates of the data in that space. This approach provides for a cheap and simple evaluation of the metrics or a cost function~\cite{muandet2017kernel}.

\textbf{Generative models} A generative auto-encoder model consists of an encoder $\mathcal{E}: P_X\sim{}X\rightarrow{}Z\sim{}P_Z$ to map the data space into latent space of codes. Latent space $Z$ has a set distribution $P_Z$. The code is then mapped to the image space $X$ with a generator (a decoder), which is a generative model $P(X\mid Z)$ deterministically mapping latent $Z$ to $\hat{X}=G(X)$. The generative auto-encoder is trained to minimise the sum of recreation cost $c(X,G(X))$ and a distance $d(Q(Z\mid X),P_Z)$ between the distribution of encoded data points $Q(Z\mid{}X)$ and the set distribution $P_Z$.

An adversarially trained GAN model lacks an encoder. Instead, after sampling a code $z\in{}Z$ and computing $G(z)$ with a generator, it uses a discriminator to answer whether the produced object comes from the $P_X$ distribution. This makes use of a minimax game.

\textbf{Kernels} We are given a~kernel $k$ in the space $Z=\R^D$ and the quality measure $\ri:Z \to [0,1]$, where higher values are better. We are going to introduce the quality measure for curves. For a~review of kernel methods see~\cite{muandet2017kernel}.

A~kernel $k:Z \times Z \to \R$ is a~function, such that there exists an embedding $\Phi$ into a Hilbert space $H$ with scalar product $\il{,}_H$ such that 
\begin{linenomath*}
	\begin{equation*}
		k(x,y)=\il{\Phi(x),\Phi(y)}_H.
	\end{equation*}
\end{linenomath*}

Let us first discuss some basic examples of kernels. A trivial kernel is linear when $H=Z$ and the embedding $\Phi$ is an identity. Thus for the linear kernel, we should use the standard distance on~$Z$. Another, most commonly encountered, is the Gaussian kernel given by the embedding  $Z \ni z \to \nor(0,\alpha I)$ into the space of square-integrable functions on $Z$, for $\alpha>0$.

In some cases, instead of finding a formula for the kernel, the embedding $\Phi$ is constructed by a~network. A typical example is the MMD GAN~\cite{li2017mmd}, where the kernel is designed so that it can play the role of the discriminator.

From our point of view, an important case is one of the generative models. Then $Z$ can be identified with the latent space and $\Phi$ with the generator. Let $X$ be a~sample $X$ in $H$ (input space), and suppose that we can generate points from the distribution by taking
$\Phi(z)$ for $z\sim P_Z$,
where $P_Z$ is the prior distribution of a generative model, typically a Normal distribution.
For a~generative model, the data points are decoded from the corresponding points in $Z$. On the other hand, the calculated distance is based on the original space's objects $H$ reconstructed from $Z$.  

Let $\gamma:[0,T] \to Z$ be a~given curve.  Additionally $\gamma$ be naturally parameterised with respect to kernel $k$, i.e.:
\begin{equation}
	\|(\Phi\gamma)'(t)\|_H=1 \text{ for }t \in [0,T].
\end{equation}

\textbf{Quality measure of a~curve} Now we will derive a~formula for the quality of a~curve $\gamma:[0,T] \to Z$. We aim to simultaneously minimise the curve's length (as computed by the kernel) and maximise the quality measure value $\ri(z)$ for each $z$ on the curve.

Let $\gamma$ be naturally parameterised, i.e.
\begin{linenomath*}
	\begin{equation*}
		\|(\Phi\gamma)'(t)\|_H=1 \text{ for } t \in [0,T].
	\end{equation*}
\end{linenomath*}
Let us fix a~small step-size $h$, and obtain the sequence of points
\begin{linenomath*}
	\begin{equation*}
		\gamma(0),\gamma(h),\ldots,\gamma(n\, h) \text{ where }n=\lfloor T/h \rfloor.
	\end{equation*}
\end{linenomath*}

We shall now take $\ri(z)$ as the probability that $z$ satisfies the requirements that were set. 
In other words, checking for all the above points $\gamma(k\,h)$, we end up with the probability of accepting the whole sequence
\begin{linenomath*}
	\begin{equation*}
		\ri(\gamma(0)) \cdot \ri(\gamma(h)) \cdot  \ldots \cdot \ri(\gamma(nh))=\exp\big[\textstyle\sum_{i=0}^n\log \ri(\gamma(i\,h))\big].
	\end{equation*}
\end{linenomath*}
By proceeding with $h \to 0$ and normalising the integral by $h$ we obtain the value
\begin{linenomath*}
	\begin{equation*}
		\exp\left(\int_0^T \log \ri(\gamma(t)) \,dt\right).
	\end{equation*}
\end{linenomath*}
Thus, for an arbitrary and not necessarily naturally parameterised curve, we obtain the following definition (we can reduce this formula to the case where $\gamma$ is parameterised by the interval $[0,1]$ without loss of generality).
\begin{definition}\label{def:quality_measure}
	For an arbitrary curve $\gamma:[0,1] \to Z$ we define the 
	quality measure of $\gamma$ by
	\begin{equation}
		\ri(\gamma)=\exp\bigg(\int_0^1 \log \ri (\gamma(t)) \|(\Phi \gamma)'(t) \|_H \, dt\bigg) \in [0,1].
	\end{equation}
\end{definition}


\begin{remark} \label{re:1}
	Observe that if $\ri \in (0,1)$ is a constant function, then directly from definition logarithm of the quality measure of a curve is negatively proportional to its distance. Thus the question of finding curves with maximal quality measure, which is the main subject of the paper, is in this case is equivalent to finding curves with minimal distance, i.e. geodesics (in the manifold $M=\Phi(Z) \subset H$ with respect to the distance given by $\|\cdot\|_H$).
\end{remark}

\textbf{Connection with geodesics in the general case of arbitrary (non-constant) quality measure}
We shall prove below that the quality measure of a curve is equal to its length with respect to a certain Riemann structure on the latent space.
\begin{theorem} \label{th1}
	Let the Riemann structure in the latent space $Z$ be defined with a~local scalar product~$\il{,}_z$ at a~point $z$ using the following formula:
	\begin{linenomath*}
		\begin{equation*}
			\il{v,w}_z= v^T A_z w \text{ where }A_z=\log^2 (\ri(z)) d\Phi(z)^T d\Phi(z).
		\end{equation*}
	\end{linenomath*}
	Then $\ri(\gamma)=\exp(-\mathrm{L}(\gamma))$,
	where $\mathrm{L}$ denotes the length of a~path $\gamma$.
\end{theorem}
\begin{proof}
	From the \cref{def:quality_measure}, we get the formula:
	\begin{equation*}
		\begin{array}{@{}l@{}l@{}}
			-\log& \ri(\gamma) 
			=-\int_0^1\log \ri(\Phi \gamma (t))\cdot \|(\Phi \gamma)'(t)\| dt \\[0.9ex]
			&= -\int_0^1\log \ri(\gamma(t)) \cdot \sqrt{\il{(\Phi \gamma)'(t),(\Phi \gamma)'(t)}} dt
			\\[0.9ex]
			&= -\int_0^1\log \ri(\gamma(t))  \sqrt{\gamma'(t)^T [d\Phi(\gamma(t))]^T d\Phi(\gamma(t))  \gamma'(t)} dt
			\\[0.9ex]
			&=  \int_0^1  \sqrt{\log^2\ri(\gamma(t)) \gamma'(t)^T [d\Phi(\gamma(t))]^T d\Phi(\gamma(t))  \gamma'(t)} dt
			\\[0.9ex]
			&=  \int_0^1 \sqrt{\gamma'(t)^T \big(\log^2\ri(\gamma(t)) d\Phi(\gamma(t))^T d\Phi(\gamma(t))\big)\gamma'(t)} dt,
		\end{array}
	\end{equation*}
	where $d\Phi(x)$ denotes the derivative of $\Phi$ at point $x$.

	Recall that in a Riemannian manifold with a metric tensor $g$, the length of a continuously differentiable curve $\gamma\colon [0, 1]\to Z$ is defined with
	$L(\gamma)=\int_0^1 \sqrt{  g_{\gamma(t)}(\gamma'(t),\gamma'(t)) }\,dt.$ Note that in our case, metric tensor is defined with $g_z\colon (v, w) \ni T_zZ\times T_zZ\mapsto \il{v,w}_z\in\R$, where $T_zZ$ is the tangent space at $z\in Z$. 
\end{proof}

We will utilise this result in the next section to associate the search for optimal interpolation with the search of geodesics.

\section{Construction of optimal interpolations}\label{sec:optimal-interpolation}

Considering the results from the previous section, we are able to formulate the definition of a~$\ri$-optimal curve.

\begin{definition}
	Let $\ri$ be a~quality index in $Z$	and $k$ be a~given kernel and $\gamma:[0,1] \to Z$ be a curve.
	
	We call $\gamma$, such that $\gamma(0)=z_0$, $\gamma(1)=z_1$, an optimal interpolation with respect to $\ri$, if it has the maximal quality measure out of all curves joining $z_0$ with $z_1$. 
\end{definition}

\cref{th1} allows us to reformulate the problem as a~task of finding geodesics. Consequently, there apply the standard results from Riemann geometry (see~\cite[Chapter 9]{spivak1970comprehensive}). 
Due to the uniqueness of the local minima and local convexity of the functional, we can minimise the energy functional instead
\begin{equation}
	E=\tfrac{1}{2}\int_0^1 \il{d\Phi(\gamma(t))\gamma'(t),\Phi(\gamma(t))\gamma'(t)}_{\gamma(t)} dt.
\end{equation}
By applying \cref{th1}, the optimal curve $\gamma:[0,1] \to Z$, $\gamma(0)=z_0,\gamma(1)=z_1$, with respect to the quality measure minimises
\begin{equation} \label{eq:E}
	E_{z_0}^{z_1}(\gamma)=\tfrac{1}{2}\int_0^1 \log^2(\ri(\gamma(t))) \|d\Phi(\gamma(t))\gamma'(t)\|^2 dt.
\end{equation}

\textbf{Optimisation procedure}
Geodesics can be found by performing nontrivial computations involving second derivatives. In some cases, we can significantly simplify this process. To justify this claim, we shall first introduce a~formula for the discretization of the integral in the energy functional.
\begin{remark}
	In our optimisation procedure we approximate $\gamma$ using a~neural network; for more details see \textbf{Continuous interpolating curves} and Appendix.
\end{remark}

We now divide the interval $[0,1]$ into $n$ intervals $[t_i,t_{i+1}]$, where  
\begin{equation*}
	0=t_0 < t_1 < \ldots < t_n=1.
\end{equation*}
We then obtain
\begin{linenomath*}
	\begin{align*}
		2E &\approx \textstyle\sum_{i=0}^{n-1} \log^2\left(\ri(\gamma(\tfrac{t_i+t_{i+1}}{2}))\right)  \cdot\,\frac{\|\Phi(\gamma(t_{i+1}))-\Phi(\gamma(t_i))\|_H^2}{(t_{i+1}-t_i)^2} \cdot (t_{i+1}-t_i)\\
		&=\textstyle\sum_{i=0}^{n-1} \log^2\left(\ri(\gamma(\tfrac{t_i+t_{i+1}}{2}))\right) \cdot
		\frac{\|\Phi(\gamma(t_{i+1}))-\Phi(\gamma(t_i))\|_H^2}{t_{i+1}-t_i}.
	\end{align*}
\end{linenomath*}
Consequently we need to minimise
\begin{equation}\label{eq.minimise_fun}
	\mathrm{cost}(\gamma)=\textstyle\sum_{i=0}^{n-1} \log^2\left(\ri(\gamma(\tfrac{t_i+t_{i+1}}{2}))\right) \cdot
	\frac{\|\Phi(\gamma(t_{i+1}))-\Phi(\gamma(t_i))\|_H^2}{t_{i+1}-t_i}.
\end{equation}
Observe that in the above formula we explicitly use the knowledge of embedding $\Phi$. To obtain the kernel version, we apply the equality $\|a-b\|^2=\il{a,a}-2\il{a,b}+\il{b,b}$ to obtain the formula:
\begin{linenomath*}
	\begin{align}\label{eq:cost}
		& \mathrm{cost}(\gamma)=\textstyle\sum_{i=0}^{n-1} \log^2\left( \ri(\gamma(\tfrac{t_i+t_{i+1}}{2}))\right)\; \cdot \frac{k(\gamma(t_{i+1}),\gamma(t_{i+1}))-2k(\gamma(t_{i+1}),\gamma(t_{i}))+k(\gamma(t_{i}),\gamma(t_{i}))}{t_{i+1}-t_i}.
	\end{align}
\end{linenomath*}

\textbf{Dynamically changing quality measure} Now we shall discuss the situation when, additionally, both the kernel (or the embedding) and the quality measure do depend on $t$, which we refer to as \emph{time}. The time analogy is used here to illustrate a situation in environments like video processing. The introduction of a time-dependent model gives a natural connection with the geodesics in the situation of a time-dependent Riemannian metric, thus allowing a wider class of applications.

We put $\tilde Z=Z \times [0,1]$, $\tilde H=H \times \R$ and $\tilde{\gamma}:[0,1] \ni t\to(\gamma(t),t) \in \tilde Z$.
Our aim is to find optimal interpolation between point $z_0$ at time $0$ and $z_1$ at $1$. By making an analogous reasoning as above for curve $\tilde \gamma$, we obtain that the cost function becomes 
\begin{linenomath*}
	\begin{align}
		\mathrm{cost}(\tilde\gamma) &=
		\textstyle\sum \limits_{i=0}^{n-1} \log^2\left(\ri_{s_i}(\gamma(s_i))\right) \cdot
		\Big((t_{i+1}-t_i)  \frac{\|\Phi_{s_i}(\gamma(t_{i+1}))-\Phi_{s_i}(\gamma(t_i))\|_H^2}{t_{i+1}-t_i}\Big),
	\end{align}
\end{linenomath*}
where $s_i=\tfrac{t_{i+1}+t_i}{2}$, while in the case when we are given the kernel instead of $\Phi$ we get the formula
\begin{linenomath*}
	\begin{align}
		&\mathrm{cost}(\tilde\gamma) =
		\textstyle\sum \limits_{i=0}^{n-1} \log^2(\ri_{s_i}(\gamma(s_i))) \cdot
		\Big((t_{i+1}-t_i) \; +  \tfrac{k_{s_i}(\gamma(t_{i+1}),\gamma(t_{i+1}))-2k_{s_i}(\gamma(t_i),\gamma(t_{i+1}))+k_{s_i}(\gamma(t_i),\gamma(t_i))}{t_{i+1}-t_i}\Big).
		\label{eq:cost_with_time}
	\end{align}
\end{linenomath*}

\textbf{Continuous interpolating curves}
Since we consider the interpolating curves to be continuous, rather than discrete, we aim at representing them with a~neural network $f_\theta$, see \cref{fig:our_model}. For a~sorted sequence of scalars $0=t_0 < \ldots <t_n=1$, equivalent to a~single batch, $f_\theta$ returns a~sequence of latent points $f_\theta(t_i)$ which represent an optimised path between two $z_0, z_1\in{}Z$. In order to meet the conditions that $g_\theta(0)=z_0, g_\theta(1)=z_1$, a~transformation, see \cref{fig:our_model}, is performed
\begin{linenomath*}
	\begin{equation}
		\label{eq:transformation}
		g_\theta(t)=(1-t) (z_0-f_\theta(0))+t(z_1-f_\theta(1))+ f_\theta(t).
	\end{equation}
\end{linenomath*}
In \cref{alg:method_train} we present step-by-step how to train of interpolation model.



We optimise curves using neural networks and consequently use batches composed of randomly sampled series of points $(\gamma(t), t)$. Thanks to the stochastic nature of the model's optimisation and learning algorithms, our method typically avoids local minima~\cite{bottou1991stochastic}.

\begin{algorithm}
	\caption{How to train the interpolation model?}\label{alg:method_train}
	\KwData{Let $z_0,z_1\in{}Z$ be given latent points and be given pretrained decoder or generator model $\Phi$}
	\KwResult{Interpolation model networks $f_\theta:[0,1]\rightarrow{}Z$ such that $f_\theta(0)=z_0, f_\theta(1)=z_1$.}
	
	Initialise the model weights $\theta$\;
	\While{not convergent}{
		Randomly generate a sequence of time $n$ points $t_i$, such that $t_i<t_{i+1}$ and $t_0=0, t_n=1$\;
		For each $i=0..n$ compute $z_i=f_\theta(t_i)$ and use transformation  \cref{eq:transformation} to get $\tilde{z_i}$\;
		Compute $y_i=\Phi(\tilde{z_i})$
		Use $\tilde{z}, y_i$ to calculate $\mathrm{cost}(f_\theta)$ from \cref{eq:cost_with_time}\;
		Update the network parameters $\theta$ on $\mathrm{cost}$ using back-propagation
	}
\end{algorithm}

\textbf{Regularisation} In this paragraph, we would like to describe an important aspect of the search for optimal curves. Observe that if quality measure $\ri$ attains either the value $0$ or $1$, we can obtain singularities. Consequently, in all our experiments, we regularise the quality index by taking
\begin{equation}\label{eq:ri_epsilon}
	\ri_\e=\e+(1-2\e)\ri,
\end{equation}
with the default value $\e=0.1$, so that $\ri_\e\in[\e,1-\e]$.

\begin{wrapfigure}{r}{0.5\textwidth}
	\setlength{\abovecaptionskip}{5pt plus 0pt minus 2pt}
	\centering
	\includegraphics[width=0.43\textwidth]{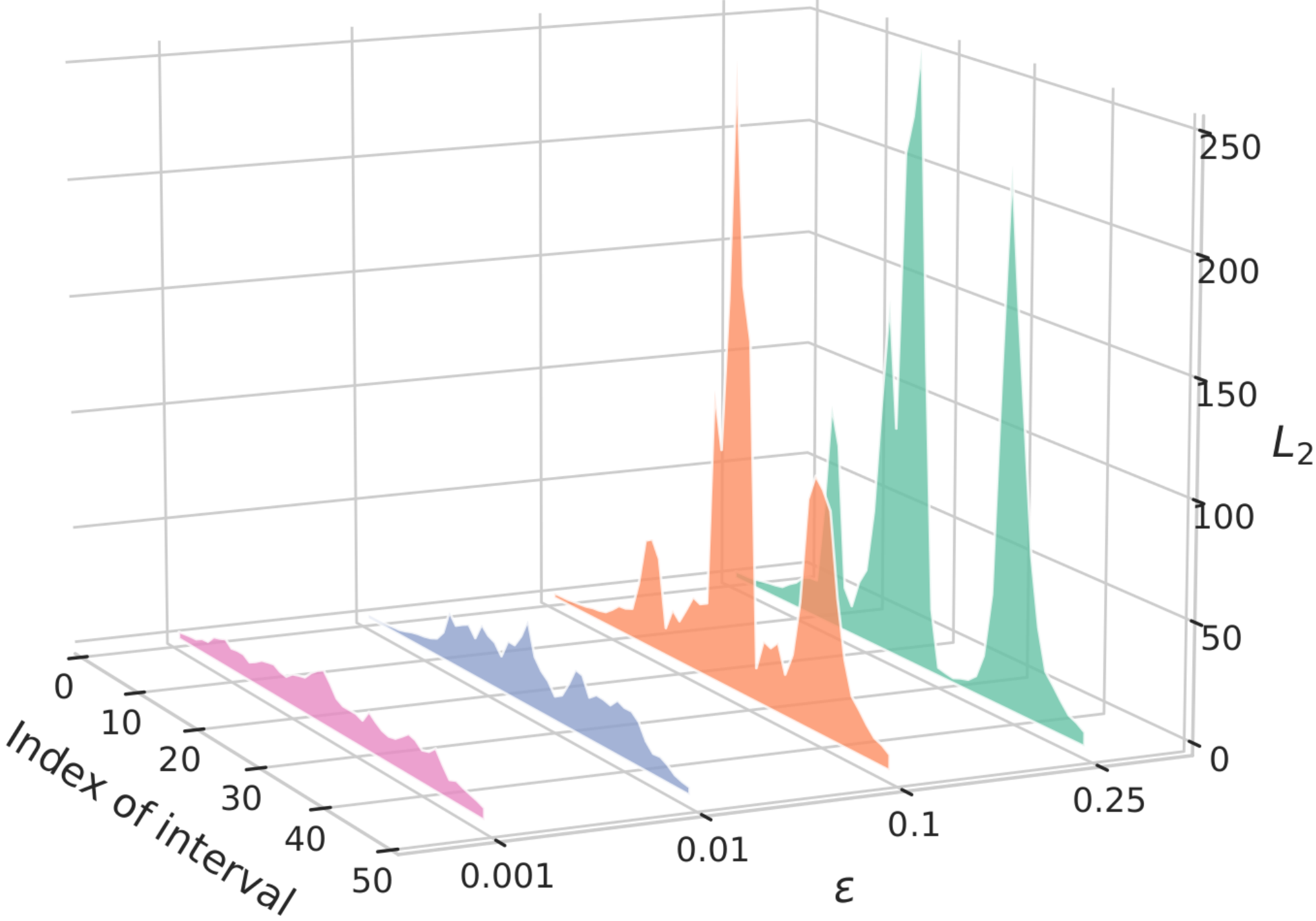}
	\caption{Squared $L_2$ distances between consecutive points for the last example path in \cref{fig:semicircle_prior} for different values of $\e$ in $\ri_\e$, see \cref{eq:ri_epsilon}. Higher $\epsilon$ values give equally spaced interpolations, while lower values result in non-equally spaced ones.} 
	\label{fig:ri_eps}
	\vspace*{-2em}
\end{wrapfigure}

\Cref{fig:ri_eps} presents squared $L_2$ distances between consecutive points in the optimised path from the last example 
from \cref{fig:semicircle_prior}. In this experiment we use different values of epsilon $\e = 0.001, 0.01, 0.1, 0.25$ for $\ri_\e$. As one can see, distances between consecutive points do not vary much for $\e$ values much larger than zero.

Setting of $\e\in[0, 0.5]$ value turns out to be important. For $\e$ close to zero, the interpolation becomes unmanageable due to the use of logarithms in the cost definition, see \cref{eq:cost}. At the same time $\e=0$ and $\ri=1$ for some domain subset may result in with logarithm of index set to zero and the interpolation to have no incentive in optimising. The $\e=0.5$ results in interpolating curve with minimal energy being equal to linear interpolation, thus not taking the probability density into account.

Higher $\e$ values result in more equally spaced interpolations, and the interpolation optimisation process is faster. However, higher $\e$ puts a different emphasis on terms of the formula optimised, i.e.~lengths start to be more important (hence the inter-point distances) than individual point realities. 

On the other hand, lower $\e$ values result in non-equally spaced points on the interpolation, which is easily visible in \cref{fig:ri_eps}, while the optimisation process is slower.

Empirically, we found $\e=0.1$ to be optimal, and therefore it is used as a default value.




	\section{Search for geodesics on manifolds}\label{sec:geodesics}
	
	In this section we will show how to apply our approach in the search of geodesics on manifolds. We start the section by 
	\begin{wrapfigure}{r}{0.4\textwidth}
		\vspace{-1em}
			\begin{center}
				\includegraphics[width=0.3\textwidth]{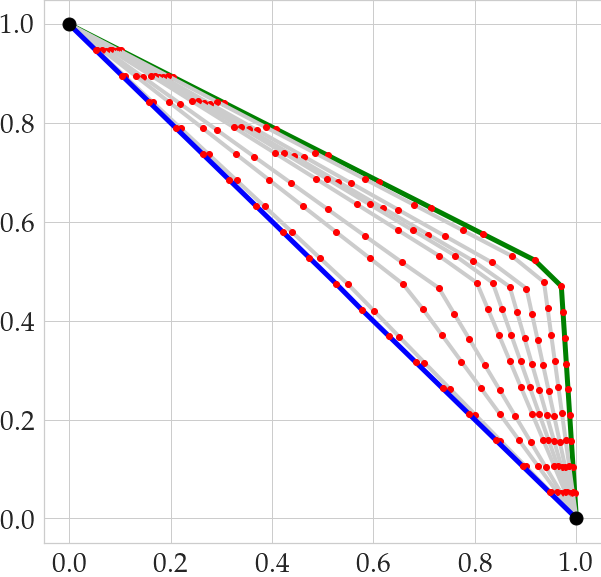}
			\end{center}
			\caption{\Cref{ex.1} results where the green (rightmost) line is the starting $\gamma$ curve and the blue one (leftmost) is the final optimised one. Lines in between them show $\gamma$ curves in subsequent optimisation stages using the cost function \cref{eq.minimise_fun}. As expected, the final $\gamma$ is a straight line connecting two endpoints.}
			\label{fig:ex1}
		\vspace*{-2.7em}
	\end{wrapfigure}
	the experiment where the manifold is given in a parametric way. Next we discuss the case when the manifold is a subset of a space (given by zero level of a function). Finally, we provide an example when the manifold is infinite dimensional with the kernel metric.
	
	By \Cref{re:1} for an arbitrary constant $\ri \in (0,1)$, the maximisation of the quality measure is equivalent to minimisation of the length of the curve. Let us show it with two simple examples. 
	
	\begin{example}\label{ex.1}
		Consider a trivial case of $Z=\R^2$. Let $\ri=1/2$ and $\Phi$ be the identity function. Take two endpoints $z_0 = (0, 1)^T$, $z_1 = (1, 0)^T$ and search for a geodesic $\gamma\colon [0, 1]\to Z$ connecting $z_1$ and $z_2$ by optimising the cost function in \cref{eq.minimise_fun}. \Cref{fig:ex1} shows several steps of $\gamma$ curve optimisation to finally get a straight segment connecting the endpoints.
	\end{example}
	
\begin{wrapfigure}{r}{0.4\textwidth}
		\vspace*{-3em}
		\begin{center}
			\includegraphics[width=0.3\textwidth]{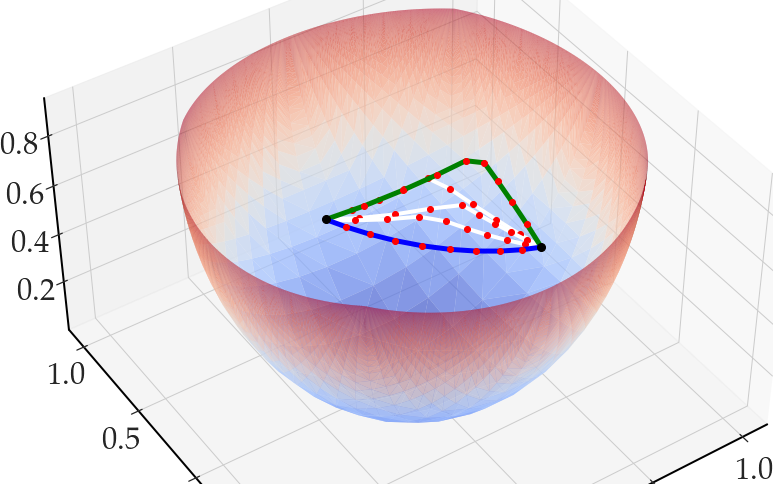}
		\end{center}
		\caption{\Cref{ex.2} results. The green line is the starting $\gamma$ curve, and the blue is the final optimised one. White lines in between present the $\gamma$ curve in subsequent stages of optimisation using cost function \cref{eq.minimise_fun}. As one can see, the final curve is the shortest line between two endpoints.}
		\label{fig:ex2}
		\vspace*{1em}
	\end{wrapfigure}

	\begin{example}\label{ex.2}
		We search now for a geodesic $\gamma\colon [0, 1]\to Z$ connecting $z_0=(0,1)^T$, $z_1=(1,0)^T$ on a half-sphere $Z=\R^2$ using cost function in \cref{eq.minimise_fun}. Let $\ri=1/2$ and 
		$\Phi(z) = (0,0,1)^T + \left((z,0)^T - (0,0,1)^T\right)/\|(z,0)^T - (0,0,1)^T\|$ 
		be given. \Cref{fig:ex2} shows the result.
	\end{example}

	Let us now proceed to a more complex case. Here we assume that, given a manifold $M$ defined in a not parametric way, but as the zero of some function. In practice we may consider an arbitrary set which is given as a zero of some function.
	Consider now the manifold $M \subset \R^n$. Since we want to find geodesics on $M$, define
	\begin{equation} \label{eq:d}
		\ri(z)=
		\begin{cases}
			\tfrac{1}{2} \text{ for } z\in M,\\
			0 \text{ otherwise.}
		\end{cases}
	\end{equation}
	Then clearly the quality measure of points outside of the manifold is zero, which means that the interpolating curve of 
	points on manifold which would not lie in $M$ is zero. Consequently, the interpolations which have nonzero quality measure have to lie in $M$.
	
	Obviously, in practice we cannot optimize with $\ri$ defined as in the above, since we need continuous functions for optimization to work. Assume therefore, that for manifold $M$
	we are given a continuous nonnegative function $d_M$ such that 
	$d_M(x)=0$ iff $x \in M$. In other words $M$ is zero level set of $d_M$. Typically, $d_M$ can be taken as a distance from the manifold
	(for a general approach to construction of such function for 2D or 3D one can look at Phi functions, see \cite{chernov2010mathematical}). Then by taking large $K>0$ we can approximate the function from \eqref{eq:d} by the following one:
	\begin{equation} \label{eq:K}
		\ri(z)=\tfrac{1}{2} \exp(-K d_M(z)).
	\end{equation}
	This means that by taking $K$ large our model will try to find curves lying in the close vicinity of the manifold $M$.
	
\begin{figure}[t]
	\centering
\begin{minipage}{.45\textwidth}
	\begin{center}
		\includegraphics[width=.95\textwidth]{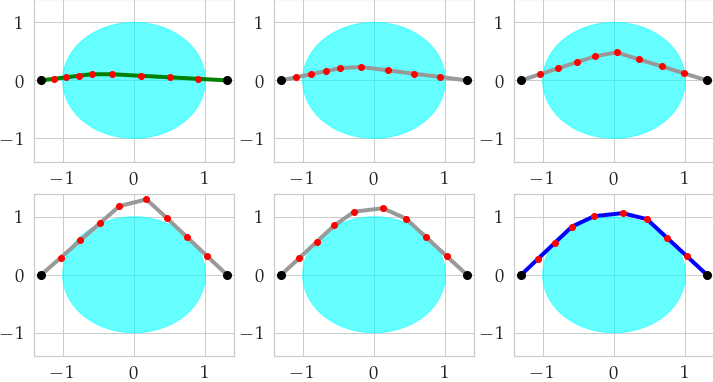}
	\end{center}
	\caption{\cref{ex.3} results. Here, the red dots on the optimised $\gamma$ path should not pass through the disk. The upper left image shows the initial shape of the $\gamma$. With subsequent optimisation steps we arrive at a solution $\gamma$ depicted in the lower right image. This final curve does not pass through the disk.}
	\label{fig:ex3}
\end{minipage}
\hspace*{3em}
\begin{minipage}{.45\textwidth}
		\begin{center}
			\includegraphics[width=0.8\textwidth]{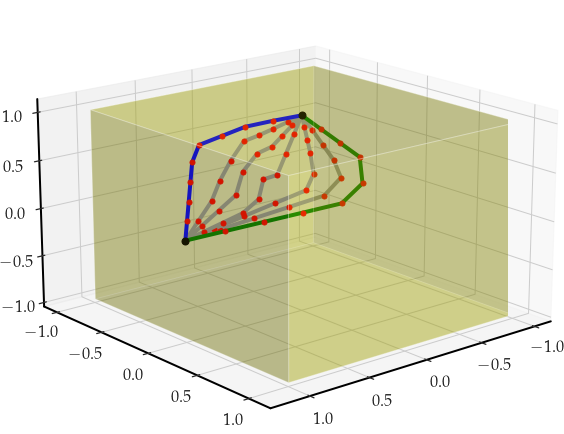}
		\end{center}
		\caption{\Cref{ex.4} results. The initial green curve enters the inside of the cube, but in the next optimisation steps the grey curves move towards two faces of the cube.}
		\label{fig:ex4}
\end{minipage}
\end{figure}

	\begin{example}\label{ex.3}
		Let $M$ be a plane with a cut-out disk with the centre $(0, 0)$ and radius $1$. Let us take a function $d_M$ that is zero on $M$ and positive outside of it by the formula
		\begin{equation*}
			\ri(z) = \tfrac{1}{2}\exp(-5 d_M(z)).
		\end{equation*}
		We apply the above with function from \eqref{eq:K} with $K=5$, and  we search for a geodesic $\gamma\colon[0, 1]\to\R^2$ connecting $z_0=(-1.3, 0)^T$ and $z_1=(1.3, 0)^T$. \Cref{fig:ex3} shows the results.
	\end{example}

	\begin{example}\label{ex.4}
		Let $M$ be cube surface. Define 
		\begin{equation*}
			d_M(z) = |1-\|z\|_{\max}|,
		\end{equation*}
		and take $K=50$ in \eqref{eq:K}.
		Using the cost function in \cref{eq.minimise_fun}, we search for a geodesic $\gamma\colon[0, 1]\to\R^3$ between $z_0=(0, 0, 1)^T$ and $z_1=(1, 0, 0)^T$. \Cref{fig:ex4} shows the results.
	\end{example}

Now we are going tp show that we can apply out approach for the search of geodesics in infinite dimensional manifold $\mathcal{M}$, consisting of probabilistic measures on $\R^n$. 
Recall that a characteristic kernel~\cite{muandet2017kernel} defines a metric on the manifold $\mathcal{M}$ consisting of probabilistic measures. Consequently, the problem of optimal interpolating between measures can be seen as an interpolation in infinite dimensional manifold $\mathcal{M}$ which, 
\begin{wrapfigure}{r}{0.45\textwidth}
	\vspace*{-1em}
	\begin{center}
		\includegraphics[width=0.43\textwidth]{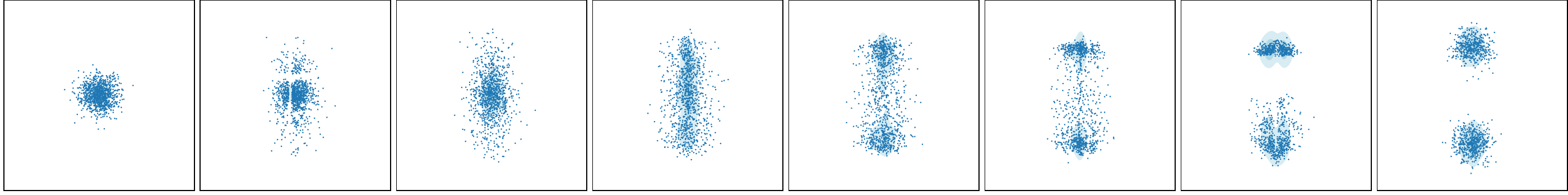}\\
		\includegraphics[width=0.43\textwidth]{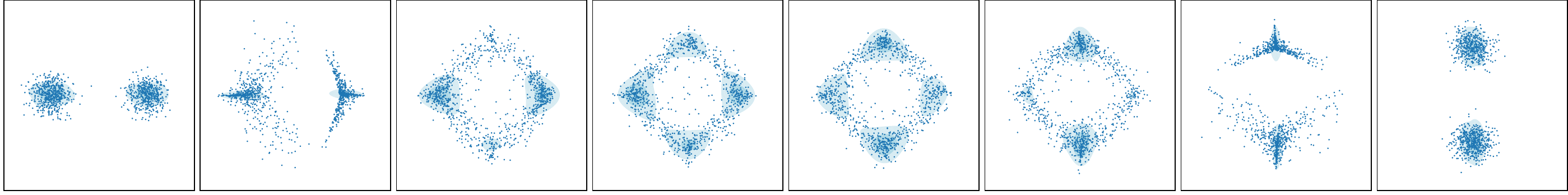}\\
		\includegraphics[width=0.43\textwidth]{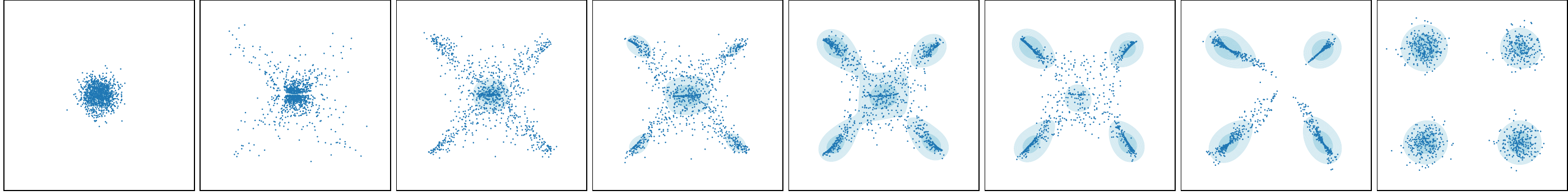}
	\end{center}
	\caption{Examples of interpolations constructed by our model in the space of measures between densities given as mixtures of Gaussians. \hspace{1em}\textbf{Top}: from $\mathcal{N}((0,0),I)$ to $\frac{1}{2}\mathcal{N}((0,-5),I)+\frac{1}{2}\mathcal{N}((0,5),I)$. \hspace{1em}\textbf{Middle}: from $\frac{1}{2}\mathcal{N}((-5,0),I)+\frac{1}{2}\mathcal{N}((5,0),I)$ to $\frac{1}{2}\mathcal{N}((0,-5),I)+\frac{1}{2}\mathcal{N}((0,5),I)$. \hspace{1em}\textbf{Bottom}: from $N((0,0),I)$ to $\frac{1}{4}\mathcal{N}((-5,-5),I)+\frac{1}{4}\mathcal{N}((-5,5),I)+\frac{1}{4}\mathcal{N}((5,-5),I)+\frac{1}{4}\mathcal{N}((5,5),I)$.}
	\label{fig:space_measure}
\end{wrapfigure}
however, can be easily visualised (we can visualise the measure by providing its sample). Thus, if we put the function $\ri$ to be an arbitrarily fixed positive constant as in the first two examples, 
our method reduces to finding geodesics. We have applied our approach, where to obtain practical implementation used samples as a estimator of measures. This interpolation can be observed in \cref{fig:space_measure}.

Suppose that we have two fixed samples $X=X^0$ and $Y=X^K$. It is our purpose to find intermediate samples between $X$ and $Y$  optimising the cost. To apply our approach, we need to be given a~kernel on this space. If a~kernel is characteristic~\cite{muandet2017kernel} (in particular, a~Gaussian kernel is characteristic), then it defines a~kernel on the space of samples of cardinality $n$~using formula
\begin{linenomath*}
	\begin{equation*}
		k(X,Y)=\textstyle\frac{1}{n^2} \sum_{i,j=1}^n k(x_i,y_j),
	\end{equation*}
\end{linenomath*}
where $X=[x_1,\ldots,x_n],Y=[y_1,\ldots,y_n] \in \R^{D \times n}$.

Since $\ri$ is constant, we find the elements of interpolating curves by minimising
\begin{linenomath*}
\begin{equation}\label{eq.cost_kernel}
	\cost (X^1,\ldots,X^{K-1}) =  
 \sum_{i=0}^{K-1} 
	\bigg( k(X^{i+1},X^{i+1}) -2k(X^{i+1},X^{i})+k(X^{i},X^{i}) \bigg) 
\end{equation}
\end{linenomath*}
and interpolation curve with minimal length. In this example, we shall use a~$D=2$--dimensional plane, $n=100$ points, and a~Gaussian kernel with $\gamma=1/8$
\begin{linenomath*}
	\begin{equation}
		\label{eq.kernel_gamma}
		k(x,y)=\exp(-\gamma \|x-y\|^2).
	\end{equation}
\end{linenomath*}
\cref{fig:space_measure} shows the results of interpolations between measures of equal cardinality. For simplicity, we take the end-points of the interpolation to be mixtures of Gaussians. We can observe that intermediate samples pass smoothly between the boundary ones and are mixtures of Gaussians too.
\Cref{sec.interpolations_measures} presents more examples of interpolations constructed by our model using other types of kernels and different values of its parameters. Nevertheless, the results (see \cref{fig:space_measure1,fig:space_measure2}) of these experiments do not show that the change of the parameter values and the type of kernel have a significant influence on the course of the interpolation.

\begin{remark}
	This shows that our solution applies an optimal transport like strategy, see e.g.~\cite{Agustsson2018,solomon2018optimal,shao2018riemannian}. We first apply some standard hypothetical curve even a linear one, and then using a neural network model we compute the distribution resulting from it. Applying the cost \cref{eq.minimise_fun} with the same network, we optimise it in the sense of the defined cost. Using an optimal transport approach, this results in the distribution of the computed curve to be consistent with the data density distribution. In the above example we end up with geodesics, since $\ri$ are constant.
\end{remark}

\section{Experiments on a generative model with arbitrary prior density}\label{sec:exp_arbitrary_secsity}

\textbf{Arbitrary prior density} As we know, in the standard generative model with prior Gaussian density, the linear or spherical interpolations are usually of satisfying quality.
This section shows how to construct interpolations if the prior density of generative model is not Gaussian.

\begin{figure}[t]
	\setlength{\abovecaptionskip}{0pt plus 0pt minus 2pt}
	\begin{center}
		\includegraphics[width=0.75\textwidth]{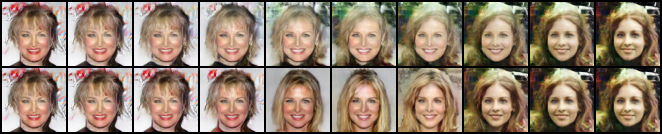}
		\includegraphics[width=0.2\textwidth]{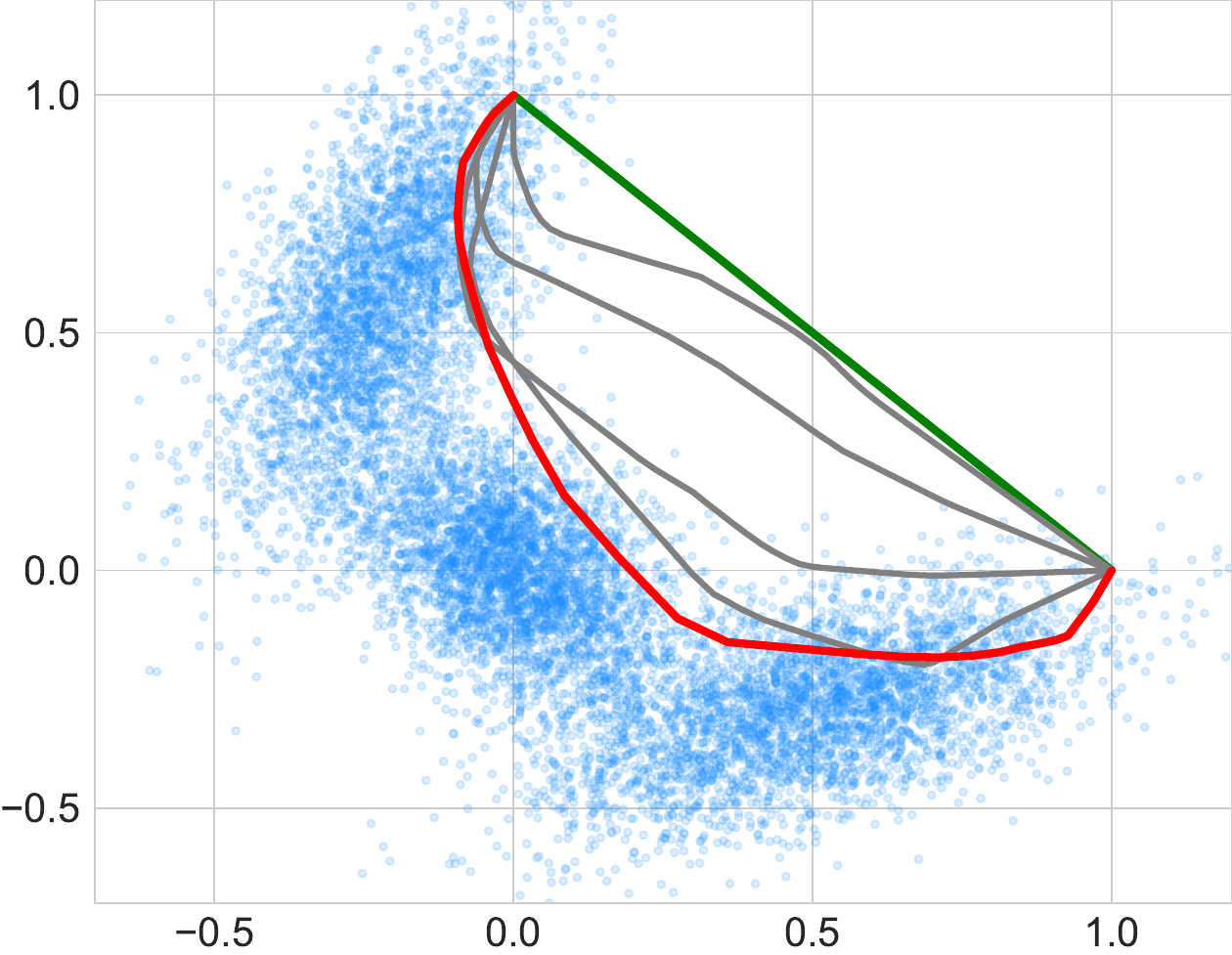} \\
		\includegraphics[width=0.75\textwidth]{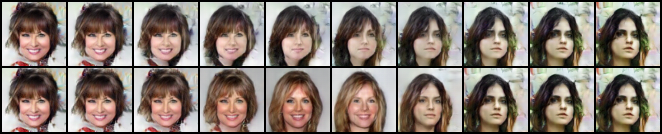}
		\includegraphics[width=0.2\textwidth]{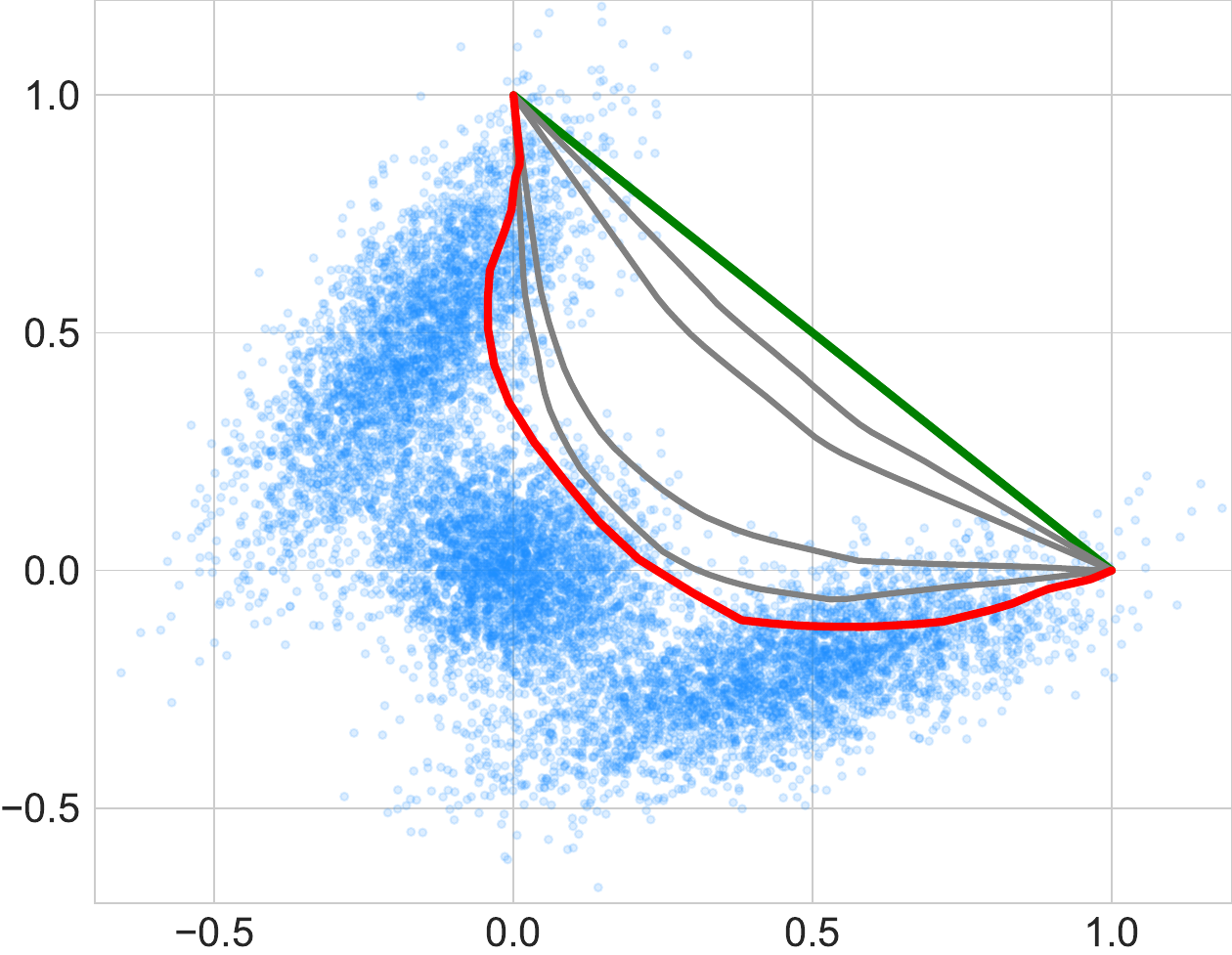} \\
		\includegraphics[width=0.455\textwidth]{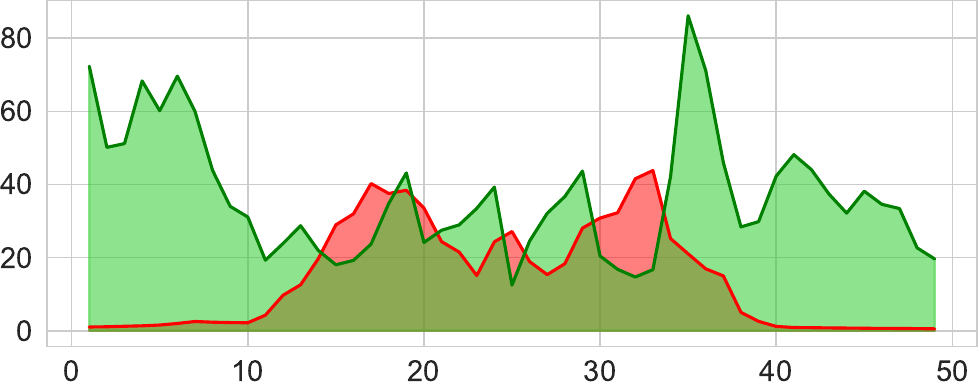}
		\qquad
		\includegraphics[width=0.455\textwidth]{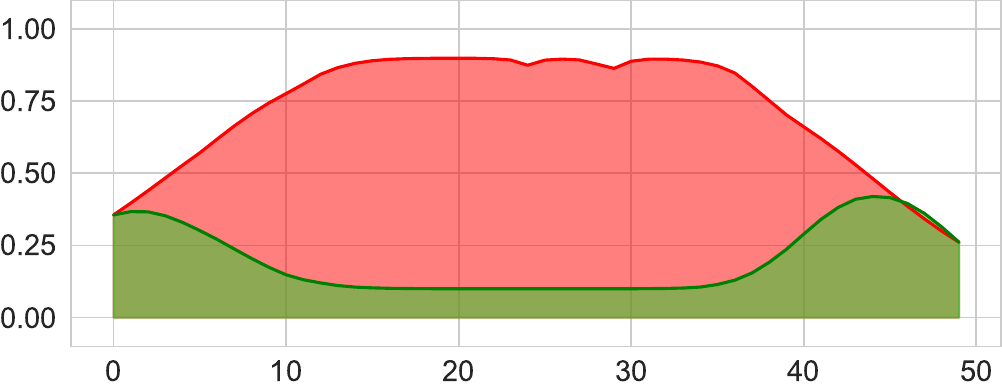} \\
	\end{center}
	\caption{\textbf{Top:} Interpolation points in a DCGAN model trained with semicircle prior latent with $D=8$. In each pair, the top row represents a linear interpolation path, while the bottom one shows the results of our optimisation procedure (for visibility, every five images are skipped in this visualisation)
		On the right-hand side, the corresponding projections (see \cref{eq:projection}) of the latent sample's density (blue dots), linear (green line), and proposed (red line) interpolations are given.
		\textbf{Bottom:} The squared $L_2$ distances between consecutive points and the quality index $\ri$ of each point in the last path above (other paths are very similar), are given (again green denotes the starting linear, and the red curves show the resulting values). It can be seen that because of the selection of endpoint images at the borders of the prior distribution, the images in between pass through an empty space resulting in low $\ri$ initial values. The proposed interpolation leads to high $\ri$ values.
	}
	\label{fig:semicircle_prior}
\end{figure}

So let us now describe our setting:
\begin{itemize}
	\setlength{\itemsep}{1pt}
	\item a~latent $Z$ with density $f$,
	\item a~generator $\Phi:Z \to X$ (for instance, $X$ contains the images from CelebA dataset), such that we can generate points from the same distribution by randomly sampling from density $f$ and transforming to the input space by $\Phi$.
\end{itemize}
We aim to construct a~quality function $\ri:Z \to [0,1]$, which would force the interpolations to proceed through high density regions. Clearly, the function $\ri$ has to depend on $f$, and higher values of $f$ should imply higher values of $\ri$. Since, in general, we do not know if $f$ is bounded, we propose to use the following normalisation:
\begin{linenomath*}
	\begin{equation*}
		\ri(z)=\mathrm{prob}(f(w) \leq f(z): w \in Z).
	\end{equation*}
\end{linenomath*}
Discuss now the situation of the case of a normal prior. Then the above formula has a~closed form given by the cumulative distribution function of the chi-square distribution. However, for the sake of analysis, one can easily obtain the following approximate formula 
\begin{linenomath*}
	\begin{equation*} \label{eq:appr}
		\ri(z) \approx \frac{1}{2}+\frac{1}{2}\mathrm{erf}\left(\sqrt{D-\tfrac{1}{2}}-\|z\|\right),
	\end{equation*}
\end{linenomath*}
where $N$ is the latent's dimensionality. We can see that the quality measure correctly identifies the latent points that lay inside the sphere $S(0,D^{1/2})$ as having the quality value of approximately $1$, and points outside the sphere as having quality measure approximately $0$.  It was shown that for a~normal distribution, all the data fit inside the sphere, whereas elements that are outside become highly distorted~\cite{LesniakSieradzkiPodolak:2019}. Observe that this behaviour is not recognised by the density itself, which has no clear change at the border of the sphere $S(0,D^{1/2})$. 

In \cref{fig:semicircle_prior} we show an interpolation path on a~specific latent distribution that is built as a~mixture of three Gaussians (called a~"semicircle"). The procedure finely fits the distribution. Standard interpolations fail in such a situation since the examples are outside of the data distribution and the variance estimates are poor in these regions resulting in high variance~\cite{arvanitidis2018}. It is therefore hard to give viable supporting qualitative measure.

\begin{figure}[t]
	\setlength{\abovecaptionskip}{0pt plus 0pt minus 2pt}
	\begin{center}
		\includegraphics[width=0.45\textwidth]{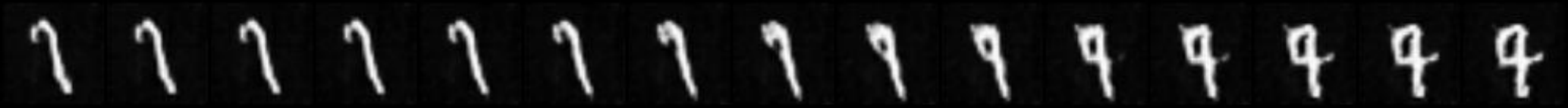}
		\quad
		\includegraphics[width=0.45\textwidth]{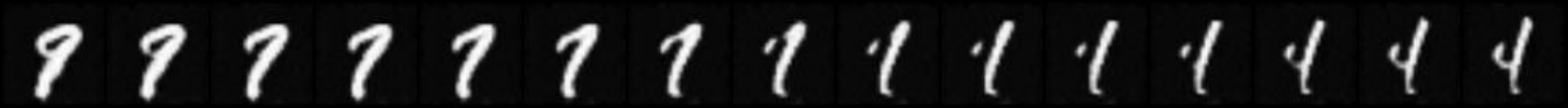}\\[-2pt]
		\includegraphics[width=0.45\textwidth]{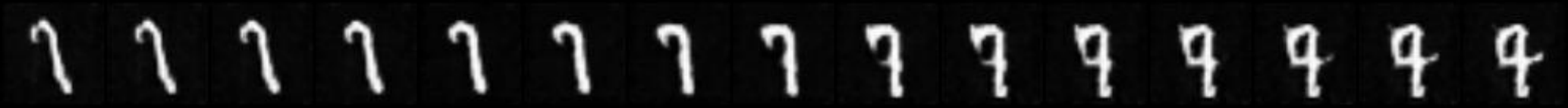}
		\quad
		\includegraphics[width=0.45\textwidth]{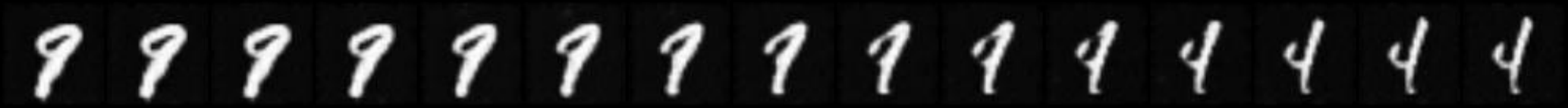}
		\\
		\includegraphics[width=0.45\textwidth]{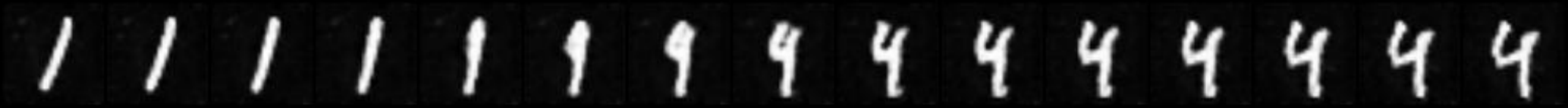}
		\quad
		\includegraphics[width=0.45\textwidth]{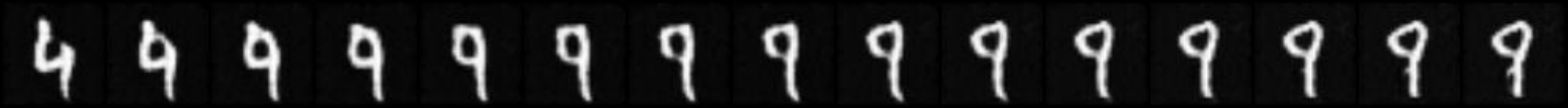}\\[-2pt]
		\includegraphics[width=0.45\textwidth]{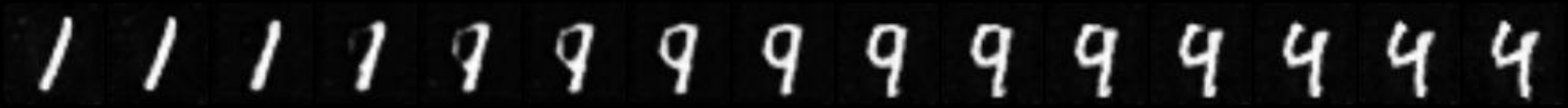}
		\quad
		\includegraphics[width=0.45\textwidth]{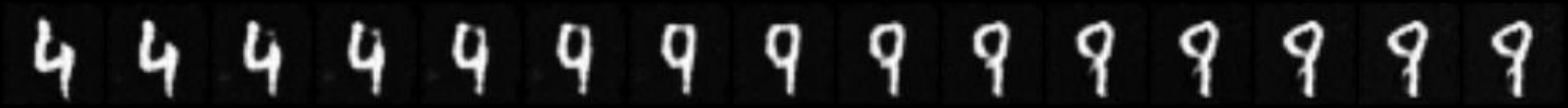}
		\\
		\includegraphics[width=0.45\textwidth]{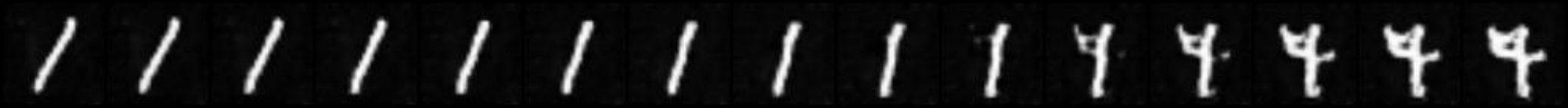}
		\quad
		\includegraphics[width=0.45\textwidth]{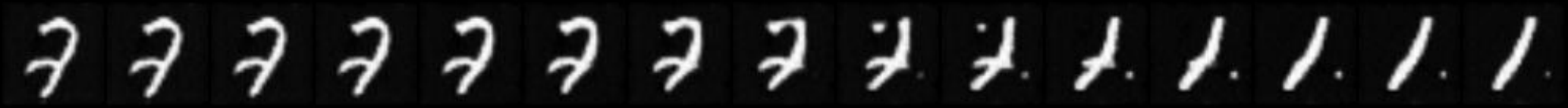}\\[-2pt]
		\includegraphics[width=0.45\textwidth]{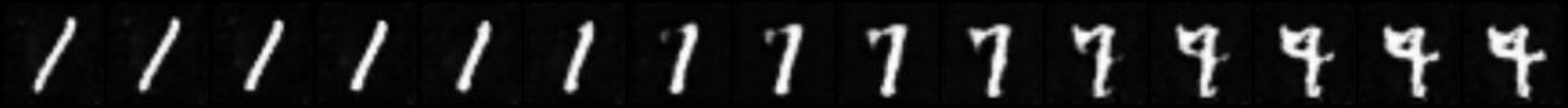}
		\quad
		\includegraphics[width=0.45\textwidth]{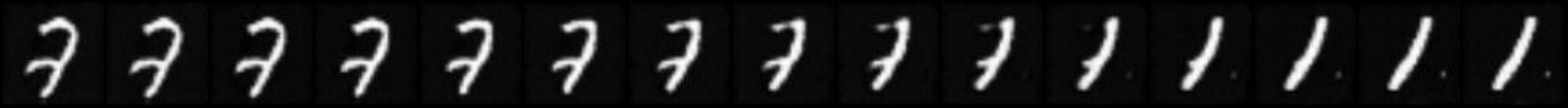}\\[2pt]
		\includegraphics[width=0.45\textwidth]{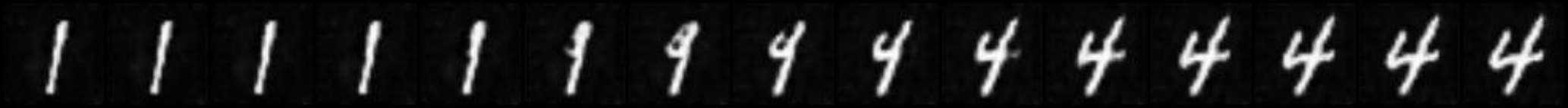}
		\quad
		\includegraphics[width=0.45\textwidth]{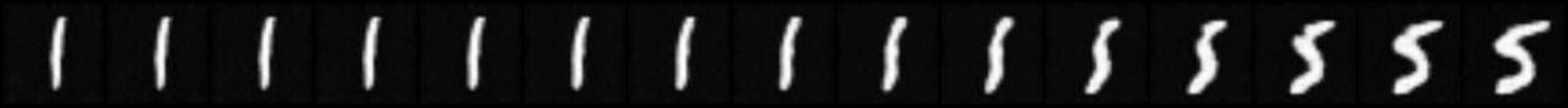}\\[-2pt]
		\includegraphics[width=0.45\textwidth]{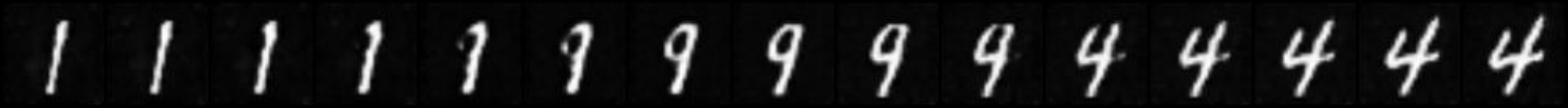}
		\quad
		\includegraphics[width=0.45\textwidth]{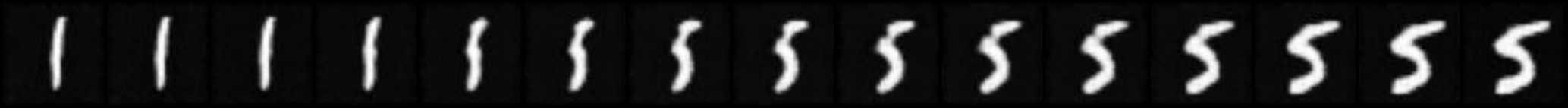}
	\end{center}
	\caption{Examples of linear and proposed interpolation paths for the MNIST dataset trained with DCGAN models, both on latent of $D=8$. \textbf{Left}:~A~semicircle latent prior model. \textbf{Right}:~A~multidimensional normal $\nor(0,I)$.}
	\label{fig:additional_paths_mnist}
\end{figure}

\begin{figure}[t]
	\begin{center}
		\includegraphics[width=0.75\textwidth]{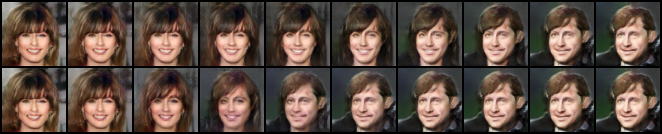}\includegraphics[width=0.2\textwidth]{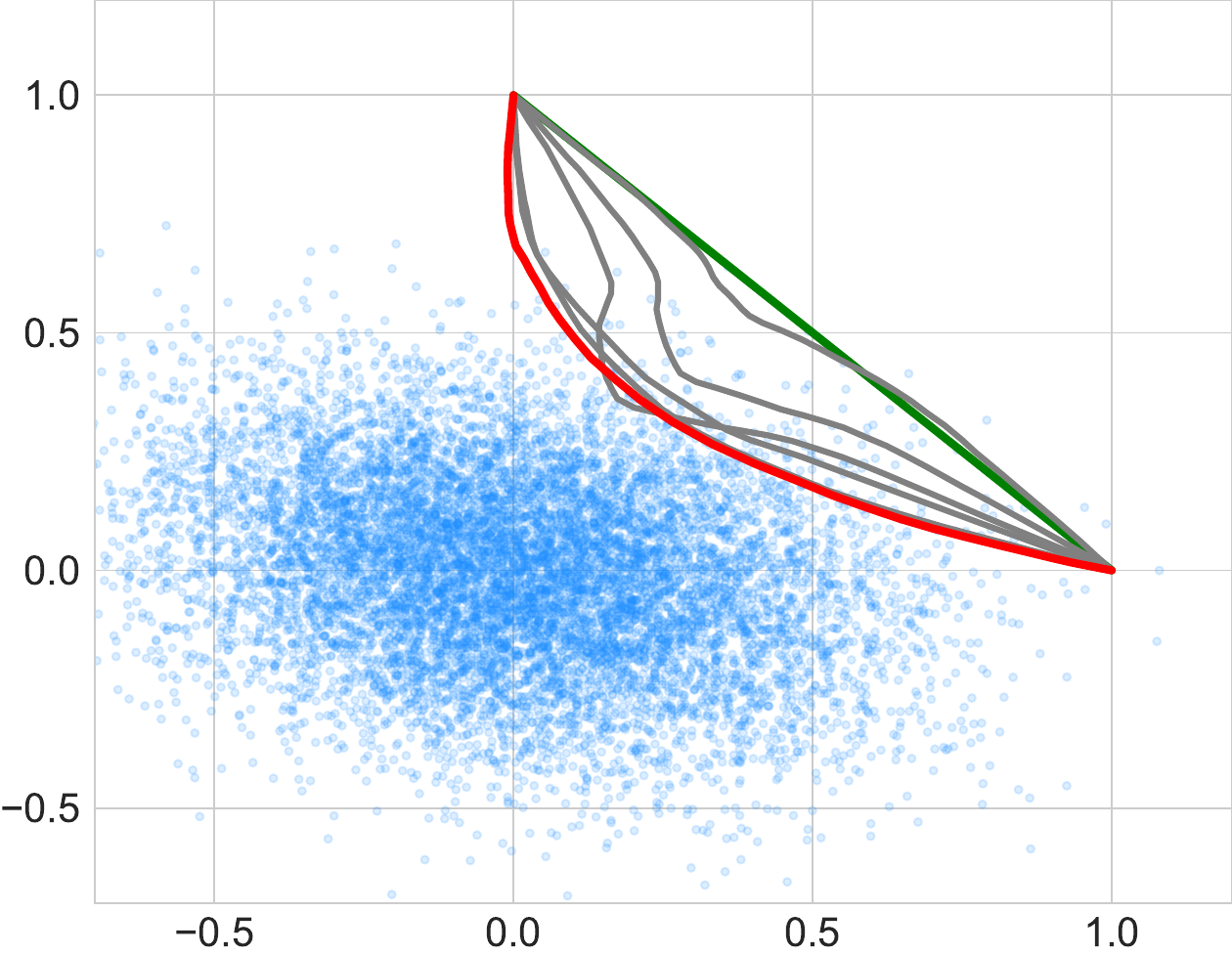}\\
		\includegraphics[width=0.75\textwidth]{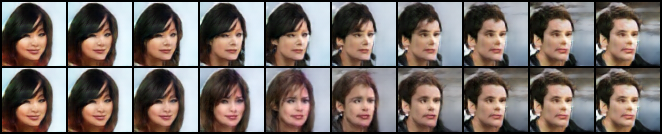}
		\includegraphics[width=0.2\textwidth]{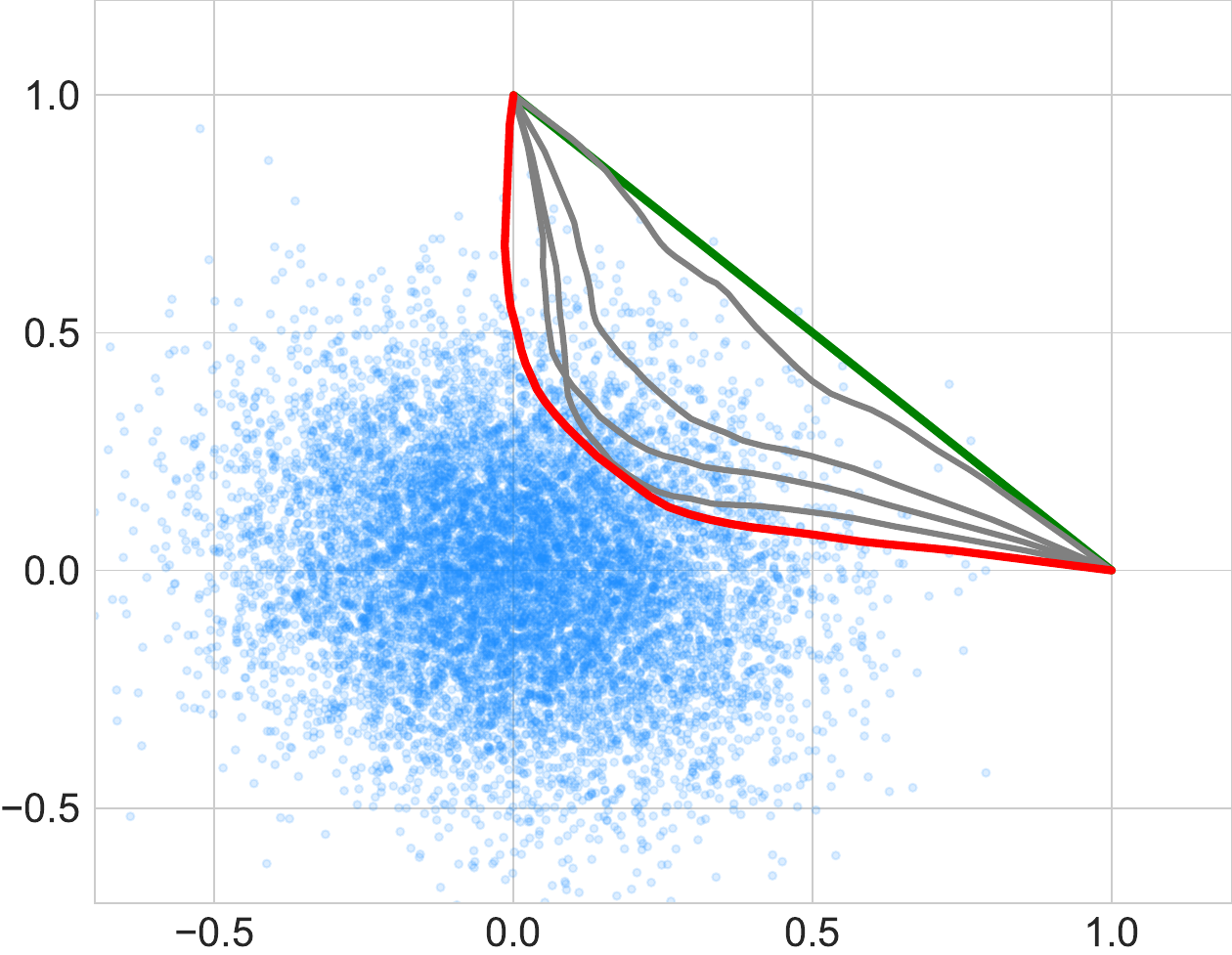} \\
		\includegraphics[width=0.455\textwidth]{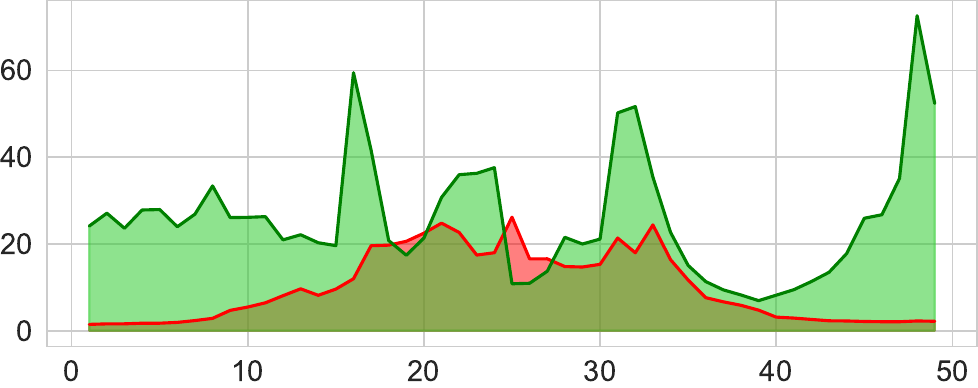}
		\qquad
		\includegraphics[width=0.455\textwidth]{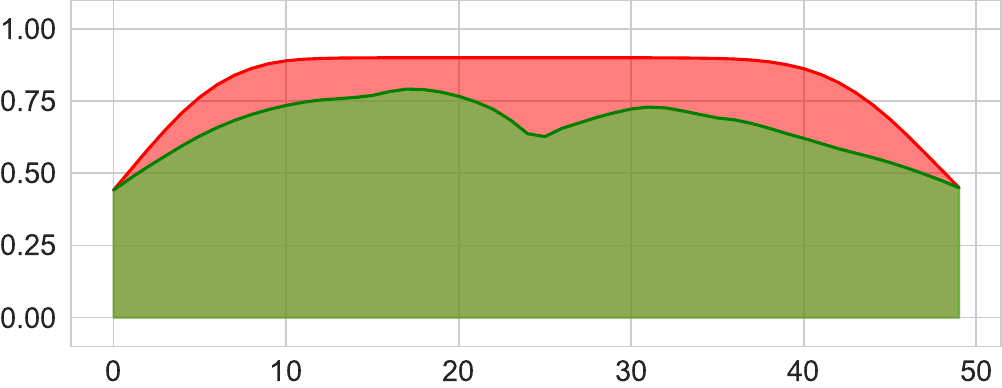} \\
	\end{center}
	\caption{\textbf{Top:} Interpolation in a DCGAN model trained with normal prior $\nor(0,I)$ with $D=20$ dimensional latent. In each pair, the upper row represents a linear interpolation path, while the lower gives results of our optimisation procedure (for visibility, every five images are skipped in this visualisation). On the right-hand side, the corresponding projections (see \cref{eq:projection}) of the latent sample's density (blue dots), linear (green line) and proposed (red line) interpolations are given. To make the figure clearer, we plotted only points in higher density areas, which is not to say that the endpoints are sampled completely from out-of-distribution data. \textbf{Bottom:} Squared $L_2$ distances between consecutive points and the quality index $\ri$ of each point in the resulting path above are given (green denotes the starting linear, and the red curves show the resulting values).
	}
	\label{fig:gaussian_prior_images}
\end{figure}

In \cref{fig:additional_paths_mnist,fig:gaussian_prior_images} we present additional results for the optimisation of the density-based quality index of an interpolation curve in a GAN model. The experiments are conducted on the CelebA and MNIST datasets, using DCGAN architectures (with latent dimensions $D=20$ for CelebA and $D=8$ for MNIST). The neural network model implementing the interpolation is initialised with weights corresponding roughly to a linear interpolation via very small weights. Each example is a sequence of 50 sorted points  $t=[t_0=0,t_1,\dots,t_{49}=1]$, with network trained using an Adam optimiser.

The results for CelebA are given in \cref{fig:gaussian_prior_images}. To better visualise path optimisation (see \cref{fig:gaussian_prior_images}, top right-hand side), we perform a projection of $k$ interpolation $z_i\in{}Z$ latent interpolation points onto $(\alpha, \beta)\in\R^2$ such that
\begin{equation}\label{eq:projection}
	\alpha\cdot z_0 + \beta\cdot z_k = \tilde{z}_i \;\text{ for }\; i = 0,\dots,k,
\end{equation} 
where $\tilde{z}_i:=X(X^{T}X)^{-1}X^{T}z_i$ are the latent space points $z_i\in Z$ and $X=[z_0, z_k]$ together with computation steps, from the linear up to the final interpolation. Notice that the proposed interpolation gives more dynamical objects. Also, the interpolation is pulled by the density. In the bottom of \cref{fig:gaussian_prior_images} progress of $L_2$ distances between consecutive images (50 in this case) and $\ri$ values are shown. It can easily be seen that interpolations become more continuous in the sense of $L_2$, at the same time becoming more realistic, i.e. more resembling examples from the data manifold.

Similar observations can be observed for a $D=8$ semicircle prior trained problems. The resulting images for CelebA, together with their path projections for semicircle prior, are shown in \cref{fig:semicircle_prior}.

We also performed corresponding experiments for the MNIST dataset trained on a $D=8$ semicircle. The resulting images for this data for the Gaussian and semicircle prior are shown in \cref{fig:additional_paths_mnist}.


\section{Finding an interpolation with additional data feature}\label{sec:exp_add_feature}
Suppose that we are given a generating model $\G$ from which we can sample from the Gaussian noise, i.e.
\begin{linenomath*}
	\begin{equation*}
		\G: Z \to \R^N,
	\end{equation*}
\end{linenomath*}
and by sampling $\G z$ for $z \sim g$ we obtain the distribution of the data $X$, where as the default $g\sim\nor(0,I)$.
Let us be interested in sampling and interpolating only some part of the data, e.g. we would like to sample and interpolate only from the numbers $1,2,7$ from the MNIST or SVHN dataset, or we are interested in persons with specified different features, e.g. colour of their eyes or hair, etc. In other words, we ask for the construction of a generative model for the subclass $Y$ of the data $X$. Let $p_Y$ denote the classifier, which returns the probability that the given element $z$ belongs to the class under consideration.

\begin{figure}[b]
	\centering
	\begin{minipage}{.46\textwidth}
		\centering
		\includegraphics[width=0.9\textwidth]{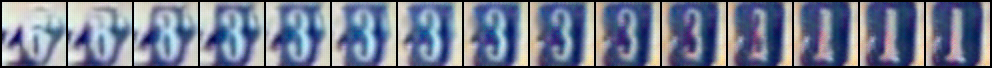}\\
		\vspace{0.1cm}
		\includegraphics[width=0.9\textwidth]{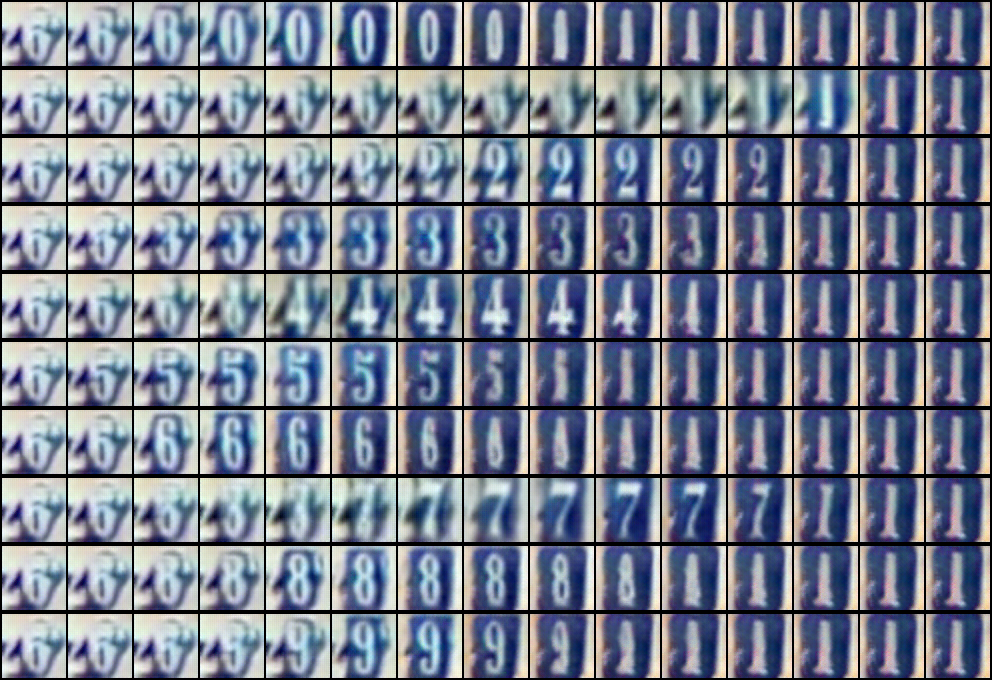}
		\caption{Examples of linear interpolation (top row) and the proposed classifier assisted results for different chosen class (each row corresponds to one chosen preferred digit). Left- and right-most points of interpolation are fixed.}
		\label{fig:svhn_all_classes}
	\end{minipage}%
	\qquad
	\begin{minipage}{0.46\textwidth}
		\centering
		\begin{tabular}{@{}c@{}c@{}}
			\includegraphics[width=\textwidth]{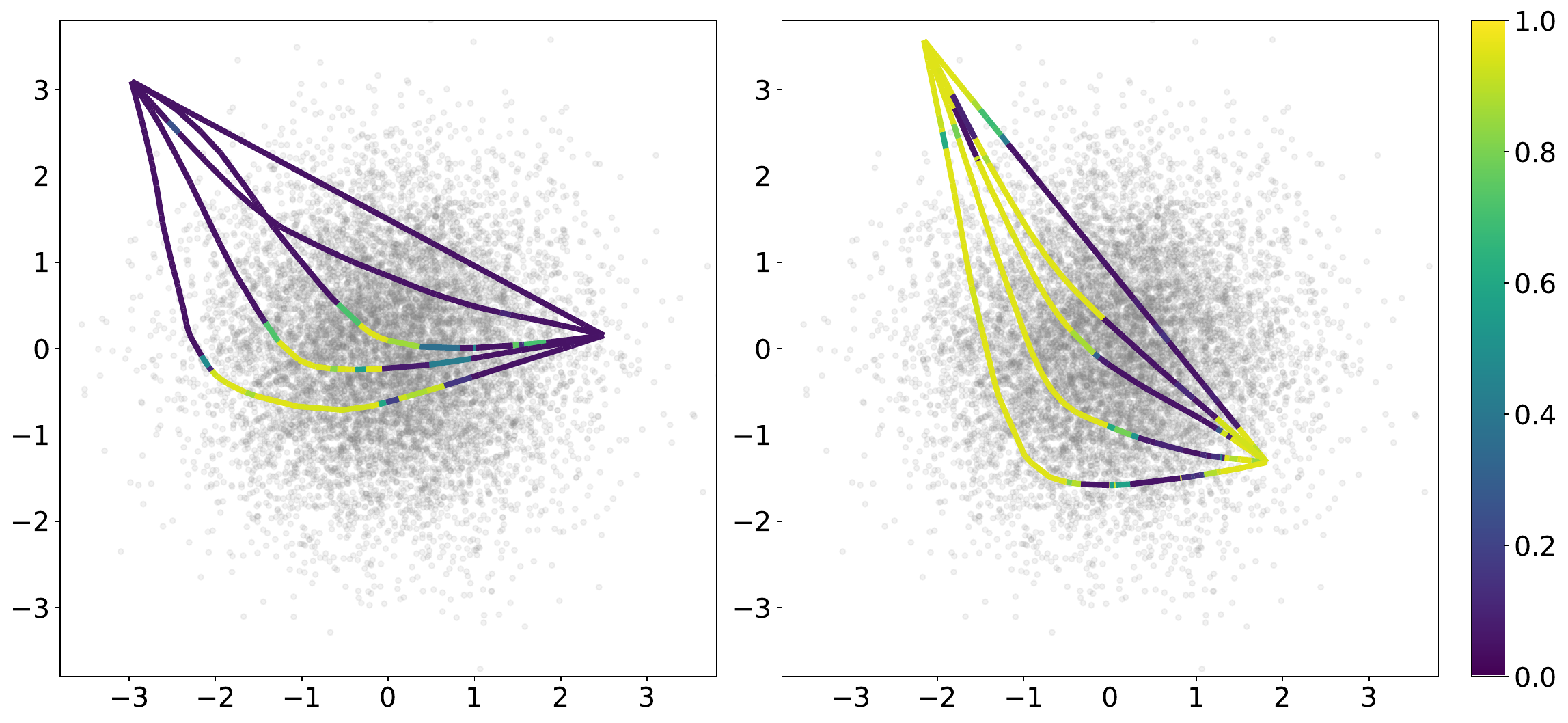}
		\end{tabular}
		\caption{Two independent interpolation mapping of SVHN images: starting from a linear (straight lines) up to the most curved. The colours denote the classifier output. Paths in the left plot favour class which does not appear in linear interpolation. In the right plot classes corresponding to boundary images are also allowed.}
		\label{fig:svhn_interpolation_3class}
	\end{minipage}
\end{figure}

Let us now describe the conditional model: the generative model which is a proper generative model on $Y$. To sample from $Y$, we can simply sample from $X$ to obtain $x$, and return $x$ as the element of $Y$ with probability $p_Y(x)$. The density of the considered model is given as the normalisation of the function
$p_Y(y) \cdot g(y)$, i.e. 
\begin{linenomath*}
	\begin{equation*}
		g_Y(y)= \frac{1}{\int p_Y(w) g(w) \, dw} p_Y(y) g(y).
	\end{equation*}
\end{linenomath*}
Observe that even if the density $g$ was Gaussian, the conditional density $g_Y$ is not. Consequently, the linear interpolation will not work properly.

An additional interesting estimation for the conditional case can be obtained if the underlying density $g$ in $Z$ is uniform on some set $U \subset Z$, i.e $g=(1/\mathrm{vol}(U))\,\1_U$,
where $\1_U$~denotes the characteristic function of the set $U$. Then one can easily observe that \begin{equation}\label{approx_loss}
	\begin{array}{l}
		\ri(z)=\frac{1}{\int_U p_Y(s) \, ds} \int_{w \in U:p_Y(w) \leq p_Y(z)} p_Y(s) \, ds \\
		\phantom{\ri(z)}
		\leq 
		\frac{\mathrm{vol}(U)}{\int p_Y(s) \, ds} p_Y(z)=C \cdot  p_Y(z),
	\end{array}
\end{equation}
for some constant $C$. Consequently, we obtain that the quality index is in this case bounded from above by a linear multiple of the classification value.

We trained a standard GAN network on the SVHN dataset~\cite{netzer2011svhn}. It was in no way optimised for our task. We trained ten separate binary classifiers recognising one of the digits. Then, using the approach described with \cref{approx_loss}, for two randomly selected $z_0,z_k\in{}Z$, the task was to find a~route that would \emph{favour} examples from a given class, at the same time minimising the energy. Results can be seen in \cref{fig:svhn_all_classes}.  Corresponding paths on projections of the latent are shown in \cref{fig:svhn_interpolation_3class}.

Let $A,B \in \R^N$ be some given images (we allow the situation $A=B$) from a~set $X \subset \R^N$, $Cl$ be a~classifier. Find an interpolating curve $\gamma:[0,1] \to Z$ such that $\Phi(\gamma(0))=A$, $\Phi(\gamma(1))=B$ and that during the interpolations $\gamma$ obtains high values of $Cl$. In a~solution for this task, we take a~generative model $\Phi:Z \to \R^N$, which generate data $X$, where the latent $Z$ has  a~prior normal density $\nor(0,I)$.

However, when we start from a point that does not have the feature recognised by $Cl$, we would force the model to arrive at it immediately. E.g. if $Cl$ classifies if the person on the image has blond hair, we would automatically enforce the appearance of this feature in the interpolation. For a remedy, we shall consider a modification
\begin{linenomath*}
	\begin{equation*}
		\ri_t(z) =1-4t(1-t) \cdot (1- Cl(\Phi(z))) \,\text{ for }\, z\in Z.
	\end{equation*}
\end{linenomath*}
Observe that $\ri_t=1$ for $t=0$ or $t=1$, which implies that we do not enforce any additional condition. Therefore we press the interpolation to gradually obtain a~given feature, where at $t=1/2$ the value is maximal, i.e. $\ri_t(z) = Cl(\Phi(z))$. 

\textbf{Pioneer model} \cref{fig:interpolation,fig:pioneer_more} shows results of experiments where the endpoints selected for interpolation corresponded to different images, which corresponds to minimisation of energy in \cref{eq:cost}. \Cref{tab:accuracy_stats_diff_ends} shows the mean ratio of images belonging to a specific class on the interpolation path, as computed by the classifier. In order to demonstrate the capabilities of our method, endpoints were selected randomly \emph{not} to have an attribute optimised.

\begin{table}[htb]
	\caption{Mean ratio of CelebA images belonging to a specific class on a interpolation path, as computed by the classifier and using a Pioneer network. Interpolation path is optimized to consist it (class True) or not to consist it (class False). For comparison, we present results for baseline spherical interpolation. Endpoints were selected randomly \emph{not} to have an attribute optimised.}
	\centering
	\begin{tabular}{lcccccc}
		\toprule
		& \multicolumn{2}{l}{Black Hair} & \multicolumn{2}{c}{Blond Hair} & \multicolumn{2}{c}{Eyeglasses} \\
		\cmidrule(lr){2-3} \cmidrule(lr){4-5} \cmidrule(lr){6-7} 
		class & False & True & False & True & False & True\\
		\midrule
		
		spherical & 0.02 & 0.00 & 0.01 & 0.00 & 0.01 & 0.00 \\
		our & 0.94 & 0.89 & 0.94 & 0.89 & 0.24 & 0.18 \\[1ex]
		\toprule
		& \multicolumn{2}{c}{Male} & \multicolumn{2}{c}{Beard} & \multicolumn{2}{c}{Young} \\
		\cmidrule(lr){2-3} \cmidrule(lr){4-5} \cmidrule(lr){6-7} 
		class & False & True & False & True & False & True\\
		\midrule
		
		spherical & 0.00 & 0.00 & 0.00 & 0.00 & 0.00 & 0.00 \\
		our & 0.68 & 0.94 & 0.95 & 0.82 & 0.94 & 0.93 \\
		\bottomrule
	\end{tabular}
	\label{tab:accuracy_stats_diff_ends}
\end{table}

\begin{figure}[t]
	\centering
	\renewcommand{\arraystretch}{0.1}
	\begin{tabular}{ @{}r@{\;}c@{\;}l@{} }
		\multirow{2}{*}{\rotatebox{90}{\makebox[10pt][c]{Pioneer}}} &
		\includegraphics[width=.95\textwidth]{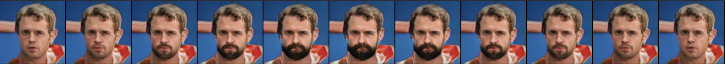}
		& \rotatebox{90}{\makebox[40pt][c]{beard}} \\
		& \includegraphics[width=.95\textwidth]{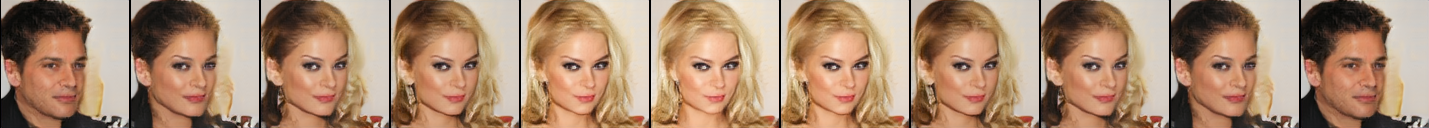}
		& \rotatebox{90}{\makebox[40pt][c]{woman}} \\
		\multirow{2}{*}{\rotatebox{90}{\makebox[5pt][c]{Glow}}} &
		\includegraphics[width=.95\textwidth]{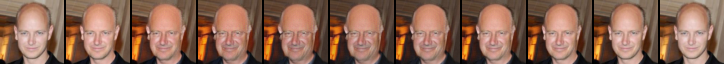}
		& \rotatebox{90}{\makebox[40pt][c]{old}} \\
		& \includegraphics[width=.95\textwidth]{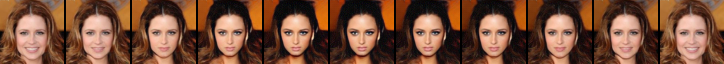}
		& \rotatebox{90}{\makebox[40pt][c]{young}} \\
	\end{tabular}
	\caption{Some results of interpolation using two different generative models. The first two rows present interpolation using a~pretrained Pioneer model and the last rows using a~pretrained Glow model between the same images trained to select images.  See more examples in \cref{app.add_results}}
	\label{fig:interpolation_the_same_image}
\end{figure}

Additionally, you can see in the table the difference of attribute constraint satisfaction between our method and a spherical interpolation. Again results from 200 paths, 50 elements each, were computed. Still, the numbers for very rare features may seem low, but the constraint of minimising both attribute presence \emph{and} the minimal path length make it hard to find such path. On the other hand, in a spherical path, the given attributes would be present only by a rare chance.

As the next example, we present some additional experiments where the defined endpoints were identical $z_0=z_1$, i.e. corresponded to the same image. In this case, an energy function with time element was used, see \cref{eq:cost_with_time}. The addition of a $(t_{i+1}-t_i)$ term pushes the curve not to stay in the same place, even though that would be the shortest path. 
\Cref{fig:interpolation_the_same_image,fig:pionier_same_endpoints_more} show paths with desired features that are the consequence of the cost function minimised. It chooses paths that has the quality needed, i.e. beard, young or old, and the short length of it. In the case of rare features, it may be impossible to fulfil both at the same time.

\textbf{Glow model} Similarly, as in the above model, we make the same experiments for the Glow~\cite{Kingma2018glow} model. The last two rows in \cref{fig:interpolation,fig:glow_diff_endpoints_more} present examples for the Glow model interpolation with the proposed framework for the case where the endpoints are different. Again some attribute optimisations are susceptible to be modified with other attribute optimisations, e.g. a~request for \emph{blond hair} or a \emph{beard} frequently changes the \emph{gender}.

On the other hand, other attributes work perfectly, as it can be seen in \cref{fig:interpolation,fig:glow_diff_endpoints_more} where the original endpoints are real-life person pictures \emph{not} taken from the CelebA, and the attributes are changed very well. It shows that our procedure makes it possible to find different paths between the same pairs of latent locations but which are focused on different attributes. We observe the same results for this model as in Pioneer model in \cref{tab:accuracy_stats_diff_ends}.

All the interpolations, be it Pioneer or the Glow base models, find very good interpolations in case of changing face view finding all the intermediate positions, e.g. see \cref{fig:interpolation,fig:pioneer_more,fig:glow_diff_endpoints_more}. Naturally, this is also the merit of the original generative models.

\cref{fig:interpolation_the_same_image,fig:glow_same_endpoints_more} show results for the Glow architecture with the same endpoints selected. The Glow generative model was in no way (just as in the case of the Pioneer) modified or optimised for the path selection. 

The optimised path images share the Glow model's benefits, although still the procedure cannot be named a perfect disentanglement, as some of the features interact. On the other hand, disentanglement was not the objective, and in the results shown, only one feature was taken into account while computing the object quality. The endpoints were selected so that they did not have the attribute that was to be optimised.

\textbf{StackGAN model} In this part, we present results on unmodified StackGAN \cite{zhang2017stackgan} model pretrained on Caltech-UCSD Birds 200 (CUB-200) dataset, which prove that our method can be successfully applied to a wide range of models. Similarly to experiments performed on Pioneer and Glow models, we select endpoints of interpolation that corresponds to different images, perform interpolation in its latent space and minimise energy defined in \cref{eq:cost}. Likewise previous models, the latent space of StackGAN models is Gaussian, despite being conditioned by text descriptions. As an image attributes to be favoured by interpolation path, we choose birds species given by classification neural network.

\begin{table}[htb]
	\caption{Mean number of CUB-200 images belonging to specific class on interpolation path, as computed by the classifier and using a StackGAN network. In order to demonstrate capabilities of our method, endpoints were selected randomly \emph{not} to have an attribute optimised.}
	\centering
	\begin{tabular}{l@{}ccc@{}}
		\toprule
		& vernilion flycatcher & american goldflinch & redhead woodpecker \\
		\midrule
		spherical & 0.01  &  0.00 & 0.02  \\
		our & 0.88 &  0.56 & 0.79 \\
		\bottomrule
	\end{tabular}
	\label{tab:accuracy_stats_diff_ends_StackGAN}
\end{table}

\begin{figure}[htb]
	\centering
	\includegraphics[width=0.95\textwidth]{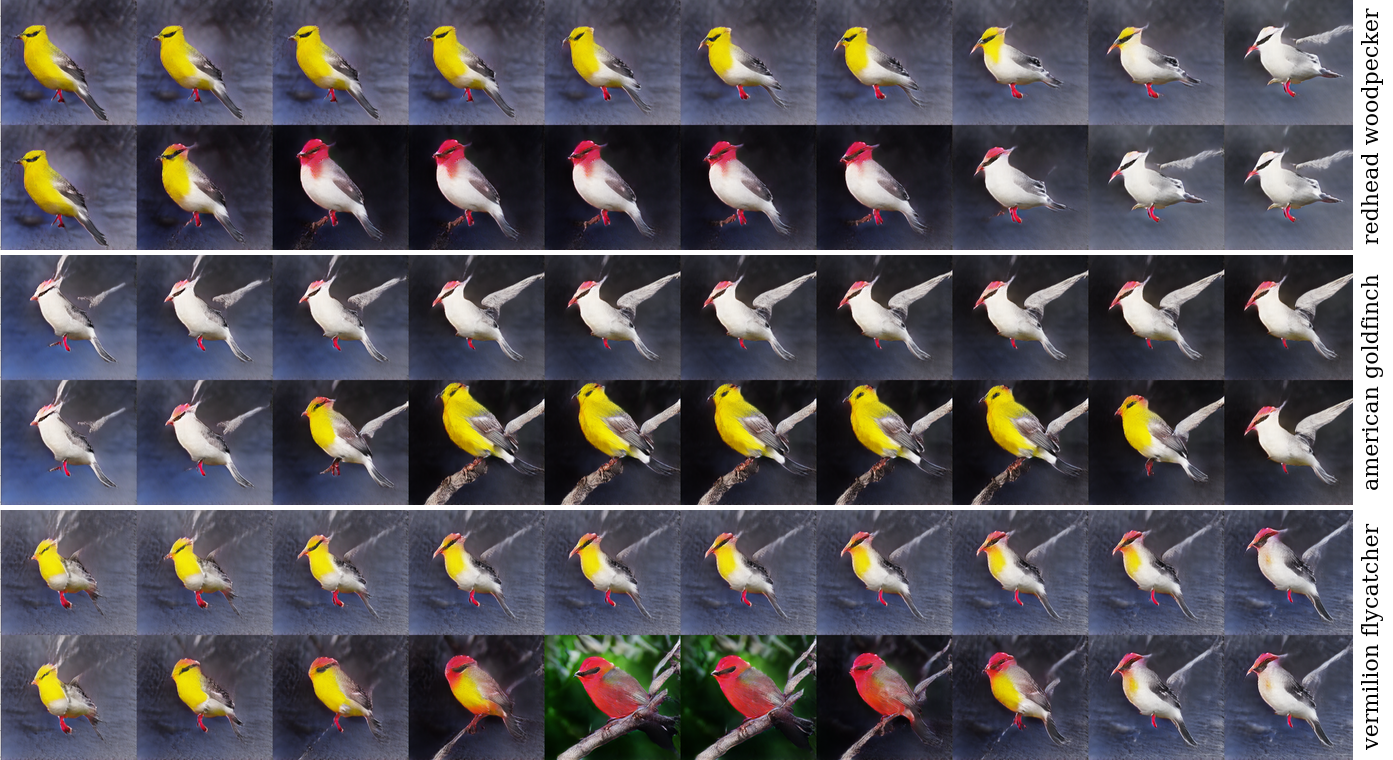}
	\caption{Examples of interpolations on StackGAN model between different images from CUB-200 dataset, which were optimised to find specific birds species. Three pairs of endpoints were chosen, then path between first pair was optimised to contain redhead woodpecker, the second to contain American goldfinch, and the third to contain vermilion flycatcher. For each example we present linear interpolation (first row) and optimised interpolation (second row).
	}
	\label{fig:StackGAN_same_endpoints}
\end{figure}

\Cref{fig:StackGAN_same_endpoints} shows examples of interpolation between three pairs of images. It compares linear interpolation with optimised path to contain specific bird species.
\cref{tab:accuracy_stats_diff_ends_StackGAN} presents the mean ratio of images belonging to a specific class on the interpolation path, as computed by the classifier. Endpoints were selected randomly with constraint \emph{not} to have an attribute to be optimised.

\bigskip

For each of these experiments in \cref{sec:exp_add_feature}, we examined different learning rates values, optimisers and weights initialisation. In all cases SGD with learning rate set to 0.001 allowed the optimisation of the model in less than 1000 iterations.

		
		


\section{Practical experiment -- automatic chemical design}\label{sec:chemical_design}

In this section, we show the application of our method to molecular optimisation. When designing novel drug candidates, the chemical space should be carefully explored in the search of molecules with improved chemical properties. Oftentimes, there exist vast subspaces of unknown compounds between the regions of potent drugs. With interpolations, we can fill these gaps and optimise the targeted properties at the same time.

In this experiment, we use two chemical properties that can be easily calculated, and so the results can be quickly validated.

\begin{itemize}
	\item \textbf{log\,P} -- the octanol-water partition coefficient describing compound hydrophobicity. We calculated log\,P using Crippen's approach implemented in the RDKit package~\cite{landrum2016rdkit}.
	\item \textbf{QED} -- quantitative estimate of drug-likeness~\cite{bickerton2012quantifying}. We use the RDKit implementation.
\end{itemize}

JT-VAE~\cite{jin2018junction} is used as a chemical generative model. The latent space of the model is continuous, but to decode molecular graphs, the decoder network uses non differentiable operations. Instead of calculating properties directly on the generated compounds, we train a classifier on the latent space to assess the data quality. For the log\,P values, we use threshold 0 to get two groups of compounds, more hydrophilic and more hydrophobic. The QED values are placed in range from 0 to 1, where higher values denote more drug like compounds. We use a threshold 0.9 for the QED optimisation. To train these binary classifiers 5000 samples from the latent space are used with the property values calculated on the decoded molecules. The classifier is a 4-layer MLP with the ReLU activation function, batch normalisation and dropout after each layer.

\begin{figure}[ht]
	\centering
	\includegraphics[width=0.9\textwidth]{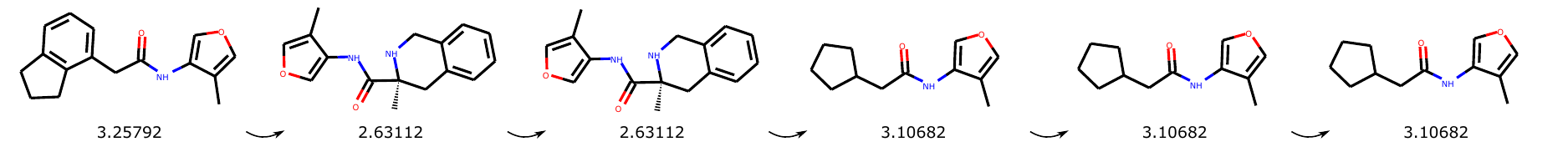}
	\includegraphics[width=0.9\textwidth]{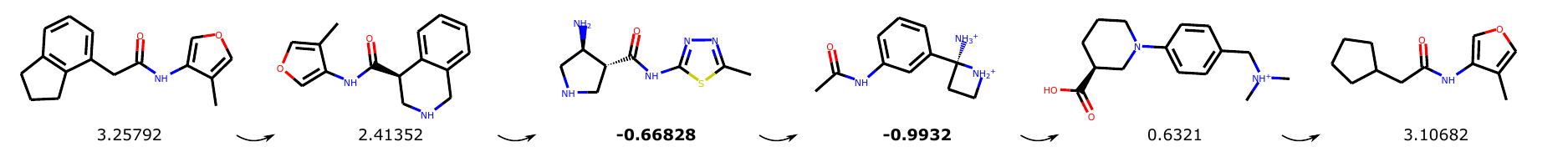}
	\caption{Examples of interpolations between two molecules. The top row presents a linear interpolation, and the bottom row depicts an interpolation that tries to minimise the log\,P value on the interpolation path. The calculated log\,P values are shown below each molecule. The values marked in bold are lower than the binarization threshold of the classifier used for optimisation (we encourage low values).}
	\label{fig:mol_logp}
\end{figure}

\begin{figure}[ht]
	\centering
	\includegraphics[width=0.91\textwidth]{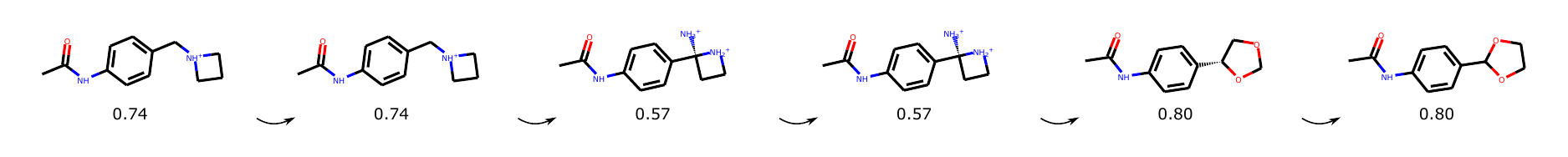}
	\includegraphics[width=0.91\textwidth]{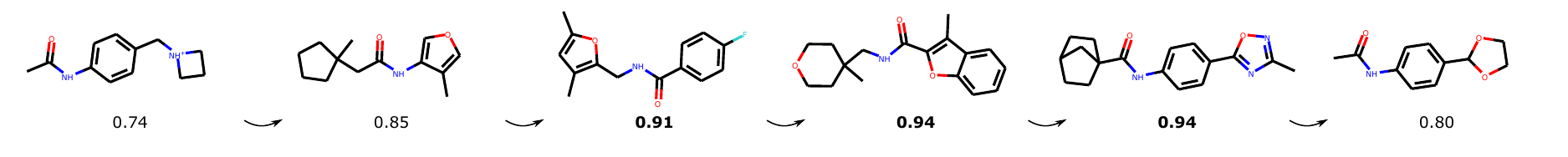}
	\caption{Examples of interpolations between two molecules. The top row presents a linear interpolation, and the bottom row depicts an interpolation that tries to maximise the QED value on the interpolation path. The calculated QED values are shown below each molecule. The values marked in bold surpass the binarization threshold of the classifier used for optimisation (we encourage high values).}
	\label{fig:mol_qed}
\end{figure}

\cref{fig:mol_logp} shows interpolation between two compounds in the JT-VAE model with the log\,P values above 3. The top row shows the linear interpolation, and the lowest log\,P value is ~2.6. Using our interpolation method (bottom row), we manage to constrain the interpolation path to go through the compounds with negative log\,P values. We also observe that the amide group linking two rings is preserved for the compounds with the lowest property values. Additionally, spherical interpolations were also tested, but the compounds decoded on the path were the same as in the case of linear interpolation.

In \cref{fig:mol_qed}, we depict the interpolation path that maximises QED in a similar manner as described above for the log\,P values. For the linear interpolation, we only observe lower QED values in the interpolation path, while in the case of optimised path, we achieve QED values much higher than those of the original compounds.

\begin{wrapfigure}{r}{0.5\textwidth}
	\vspace{-1em}
	\centering
	\includegraphics[width=0.45\textwidth]{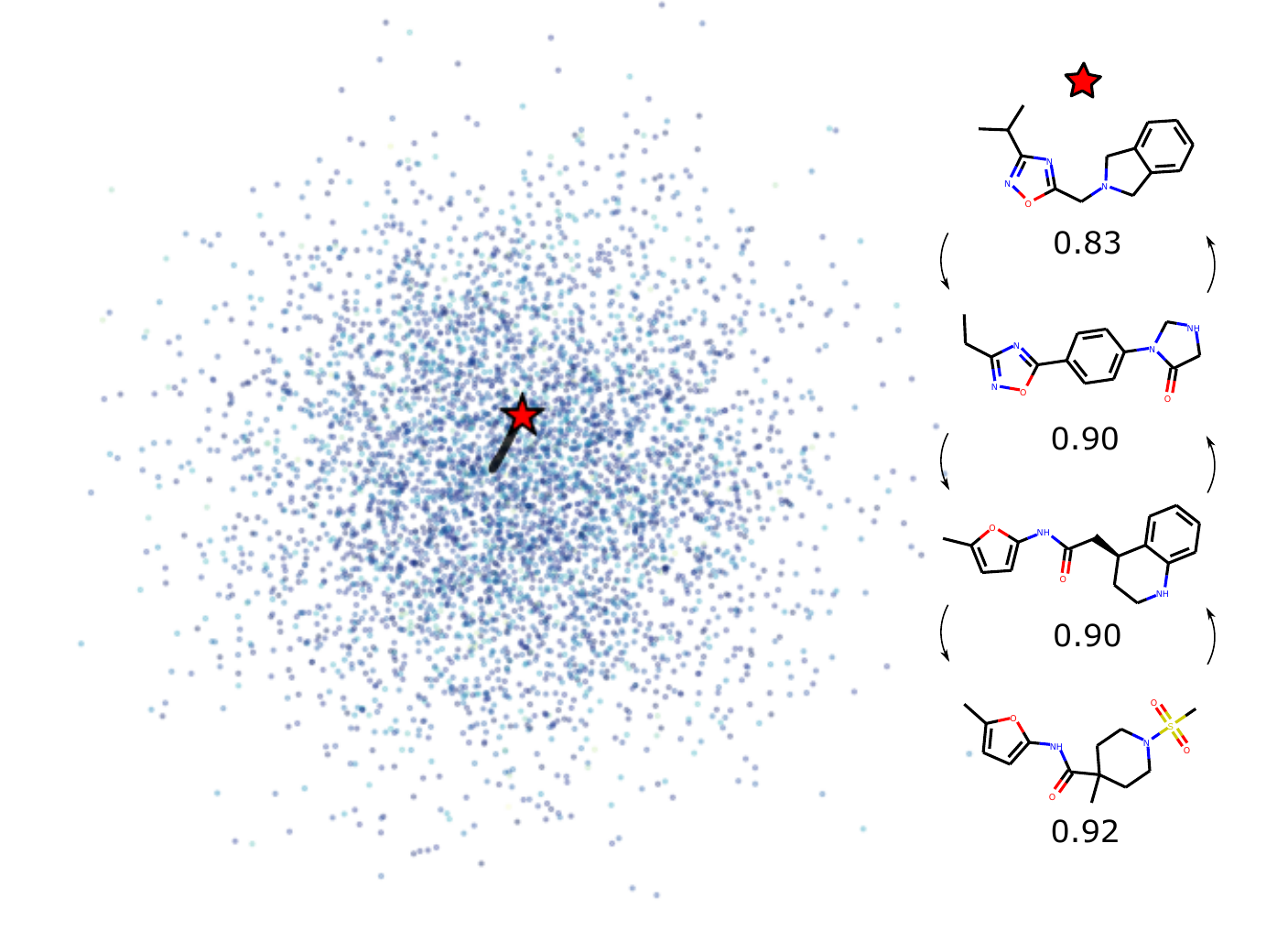}
	\caption{Optimisation of one sampled compound from the latent space (red star). The plot shows PCA of the latent space. Each point is a sample from the Gaussian distribution, and the black curve is the optimisation path. The purple colour of the points corresponds to higher QED values. On the right, the optimisation path is shown, starting with the compound marked with a red star -- QED is shown below the molecules.}
	\label{fig:self_qed}
	\vspace{-5em}
\end{wrapfigure}

\cref{fig:self_qed} also shows the optimisation path that maximises QED, but using only one compound (the first and 
last compounds on the interpolation path are the same). We observe that the optimisation path successfully leads into the region of higher QED values. Returning to the starting compounds, the same compounds are decoded as the backward path is symmetric. It is notable that all the generated molecules on the path were valid due to the use of the JT-VAE model. JT-VAE is a graph-based model that alleviates the problem of generating invalid compounds by decoding only valid substructures extracted from the organic compound databases.

\section{Conclusions}

We have studied the problem of producing a~meaningful latent space interpolation in pretrained generative models. In the paper, we claim that a~correct interpolation framework should generate paths that reveal the data hidden structure, be smooth, be composed of elements from the true data distribution, and should allow modifying the density of features during the interpolation process. The framework should fulfil it for different types of measures too.

We have proposed a~general interpolation framework that meets these requirements through a~definition of a~cost function utilising a~kernel distance and by defining the notion of a~$\ri$-optimality, which prefers to guide the path through areas of high data probability, resulting in objects close to the data manifold. Thanks to this definition, it is possible to find a~continuous path rapidly using a~dedicated neural network. This definition corresponds to geodesics search in a~Riemann structure of the latent.

Such a~framework permits to modify a~path between two fixed end-points in such a way that additional features are met. This, and the whole framework definition, allows for path generation for already trained models. In the experiments, we have used available implementations of the Pioneer network without changing anything in them, a Glow network, a StackGAN, or independently trained a JT-VAE for chemical purposes. The correctness of the backbone model, together with the $Q$ definition (see examples in \cref{sec:exp_arbitrary_secsity}), ensures that the model adheres to the distribution.

Our approach is general enough to work in the space of measures, where we can define a~method of finding a~path between two distributions, possibly satisfying some additional criteria. This corresponds to the case of restricting individual attributes. Irrespective of whether the backbone model has has the data inherent features disentangled, the curve found together with some condition, can find a path along which that particular feature changes almost linearly, e.g. see \cref{fig:interpolation,fig:interpolation_the_same_image,fig:StackGAN_same_endpoints,fig:mol_logp,fig:mol_qed}. The examples show, that our solution is able to do an interpolation where the endpoints are defined with a single object, e.g. image, molecule, etc., and do a "loop" in the latent space while changing some data space feature. 

Generative models that have a~latent with a~set prior distribution are ideal tools for producing new examples that come from the training set manifold. Aside from sampling the latent as is, it enables us to produce paths of examples that have additional attributes. The proposed model enables a possibility to change the distribution along the sampling path. This can be useful in finding new molecules in cheminformatics, in particular when all found molecules need to be synthesizable, routes for robots that avoid obstacles, finding paths of activations in a~human brain. 

\section*{Acknowledgements}
The work of \L{}ukasz Struski, Micha\l{} Sadowski and Jacek Tabor was carried out within the Team-Net program of the Foundation for Polish Science co-financed by the European Union under the European Regional Development Fund (grant no. POIR.04.04.00-00-14DE/18-00).
The research of Tomasz Danel and Igor T.~Podolak has been supported by a grant from the Priority Research Area (DigiWorld, “Stability of auto-encoder representations of molecules) under the Strategic Programme Excellence Initiative at the Jagiellonian University.

\bigskip

\noindent\rule{\textwidth}{0.4pt}
\begin{center}
	\textbf{\Large Appendix}
	\noindent\rule{\textwidth}{0.4pt}
\end{center}

\medskip

\numberwithin{figure}{section}
\crefalias{section}{appsec}

\section{Interpolation model architecture}\label{app:interpolation_model}
As it was described in the main paper, the principal objective of our research was to determine a methodology (framework) for finding an optimal path between $z_0,z_1\in{}Z$ latent points such that the latent is build with independent generative models (in the experiments we have used DCGAN~\cite{radford2015unsupervised}, the Pioneer network~\cite{heljakka2018pioneer}, and Glow~\cite{Kingma2018glow} networks) with the ability to keep track of individual attributes, or a JT-VAE~\cite{jin2018junction}. This approach does not need any direct modifications of the generative models themselves -- actually we have used pretrained models that were available.

At the same time we wanted to have a time-dependent object quality measure, that would be related to the place on the parameterised path $\gamma(t)$ such that $\gamma(0)=z_0, \gamma(1)=z_1$, even allowing for the case $\gamma(0)=z_0=z_1=\gamma(1)$, i.e. the case of a "loop" exiting and returning from the same object, but prohibiting the case where the $\gamma$ path would be $0$-dimensional. This case allows for, e.g. such modifications of an image that some features of it are changed for only some time, where time is identified with parameter $t$.

To facilitate these goals, we have, in terms of kernels, defined the appropriate cost functions. In order to make it undemanding and fast to compute, we have proposed to use specially designed, although straightforward in design, neural networks that would optimise the energy functions \cref{eq:cost,eq:cost_with_time}.

The interpolation model networks implement a function $g_\theta:[0,1]\rightarrow{}Z$, i.e. the $\gamma(t)$ path. The network is input a sequence of time $n$ points $t_i$, such that $t_i<t_{i+1}$ and $t_0=0, t_n=1$. This whole sequence is treated as a batch in network learning and the appropriate energy is computed for the whole sequence. The path finding is equivalent to network training with network parameters $\theta$ being the parameters of $\gamma(t;\theta)$. In order to compute the quality of individual points, the network needs to have the access to the decoder (or generator in case of GANs) $\Phi$ of the original generative model. At the same time, to fulfil the constraint that generated endpoints are identical to input $z_0,z_1$, we apply a~simple affine transformation at the end of processing
\begin{linenomath*}
	\begin{equation*}
		g_\theta(t)=(1-t)(z_0-f_\theta(0))+t(z_1-f_\theta(1))+f_\theta(t),
	\end{equation*}
\end{linenomath*}
where $f_\theta$ are the outputs of the actual interpolation neural network, and $g_\theta$ represents the whole model.

Since the energy functions to be minimised take into account both the quality of the corresponding generated object $\Phi(\gamma(t))$ and the distances between objects, it may be impossible to define which $(t_i)$ sequence is optimum. In the experiments, we have used sequences that were drawn from a uniform distribution of scalars from $[0,1]$ segment.

Naturally, the generated latent path $\gamma$ must comply with the generative model in question. Therefore, for a Pioneer network, $g_\theta(t)$ is simply an $N$-dimensional point in the latent, but in case of the Glow network it needs to be a list of 3-dimensional matrices.
All models included a batch or layer normalisation for stable learning.

	\section{Interpolations between measures - the choice of kernel and discretization}
	\label{sec.interpolations_measures}
	
	In this section, we show a sensitivity of the optimisation probed involving objective function in \cref{eq.cost_kernel} to the choice of kernel and discretization. We consider two types of kernels: the first defined in \cref{eq.kernel_gamma} and then the inverse multi-quadratic defined with the following formula:
	\begin{equation}
		\label{eq.kernel_multiquadratic}
		k(x, y) = 1 / \sqrt{\|x - y\|^2 + c},
	\end{equation}
	where $c>0$.
	
\begin{figure}[t]
	\centering
	\begin{minipage}{.46\textwidth}
		\begin{center}
			\includegraphics[width=\textwidth]{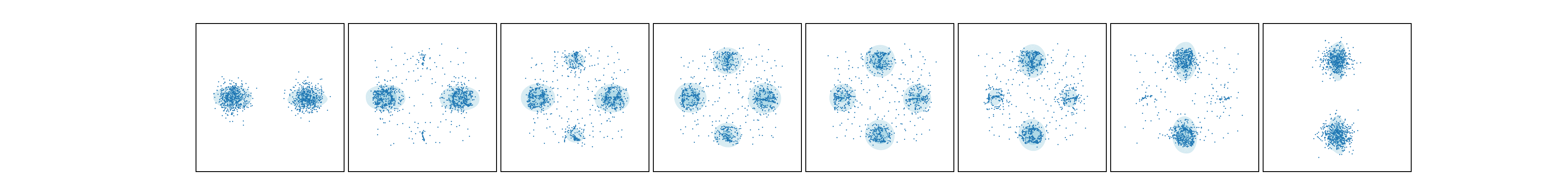} \\[-0.7em]
			\includegraphics[width=\textwidth]{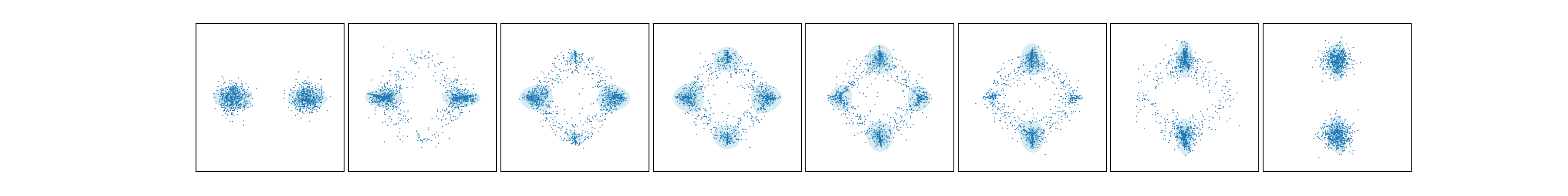} \\[-0.7em]
			\includegraphics[width=\textwidth]{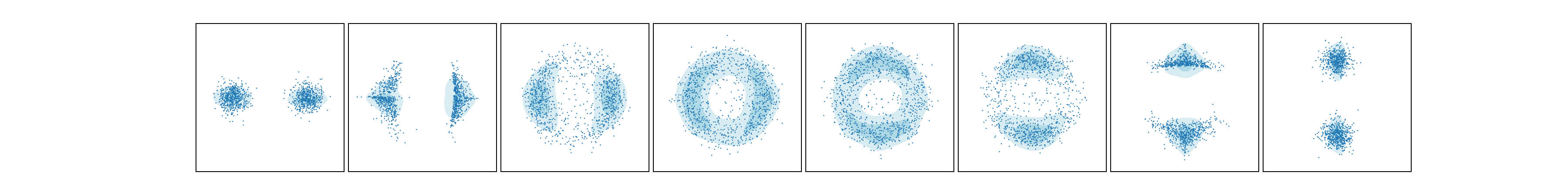} \\[0.3em] 
			\includegraphics[width=\textwidth]{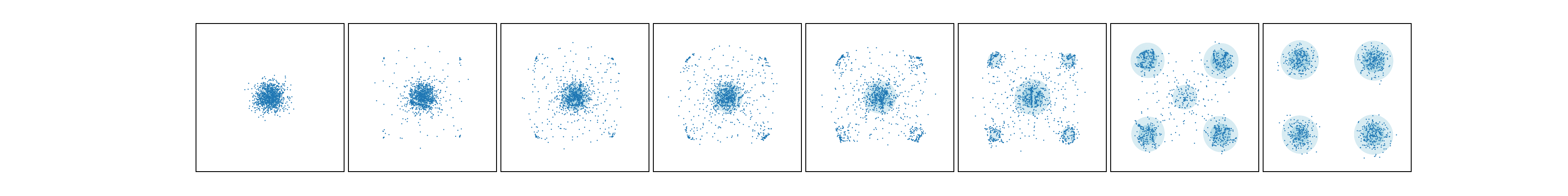} \\[-0.7em]
			\includegraphics[width=\textwidth]{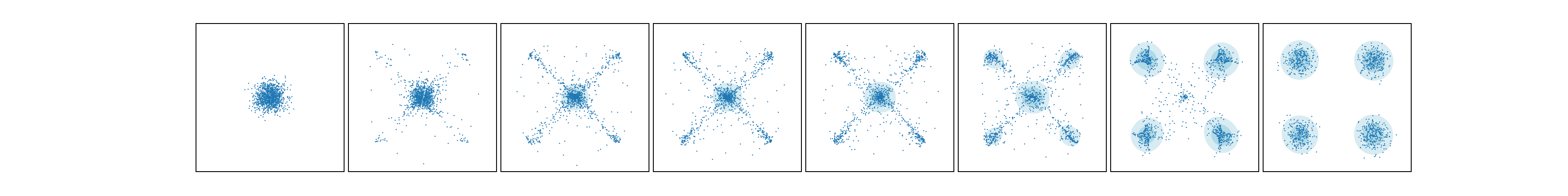} \\[-0.7em]
			\includegraphics[width=\textwidth]{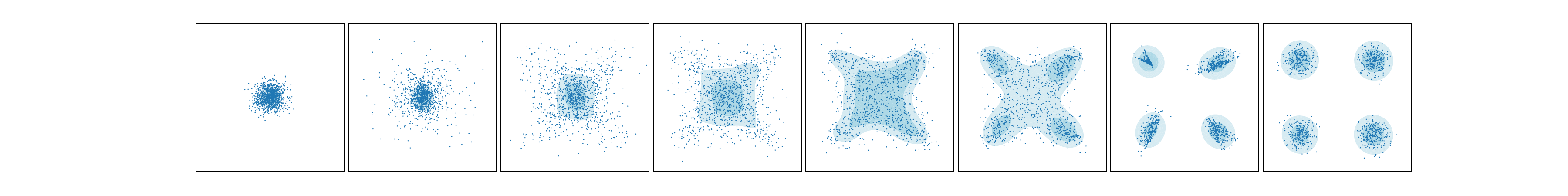}
		\end{center}
		\caption{Examples of interpolations constructed by our model in the space of measures between densities given as mixtures of Gaussians where we use as parameter $\gamma$ from \cref{eq.kernel_gamma}, we put $\gamma = 1/4, 1/16, 1/64$ (from top to bottom respectively). We show the last two examples from \cref{fig:space_measure}.}
		\label{fig:space_measure1}
	\end{minipage}%
\qquad
\begin{minipage}{0.46\textwidth}
		\begin{center}
			\includegraphics[width=\textwidth]{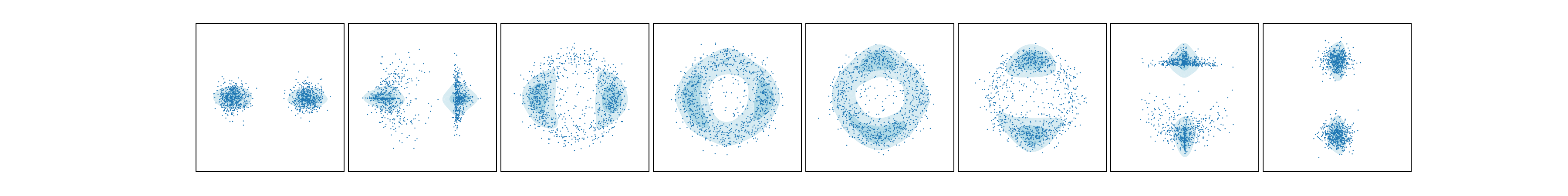} \\[-0.7em]
			\includegraphics[width=\textwidth]{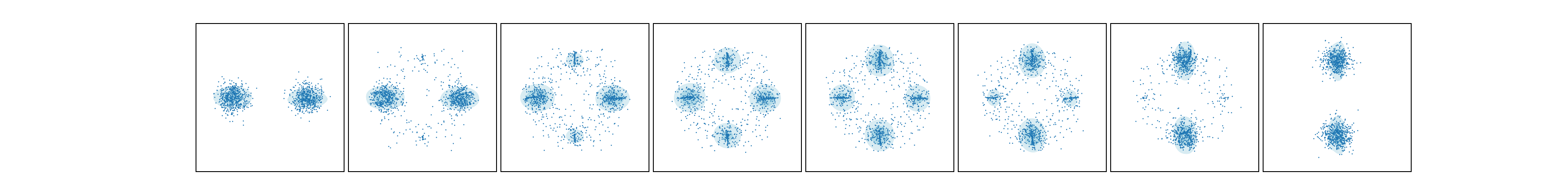} \\[-0.7em]
			\includegraphics[width=\textwidth]{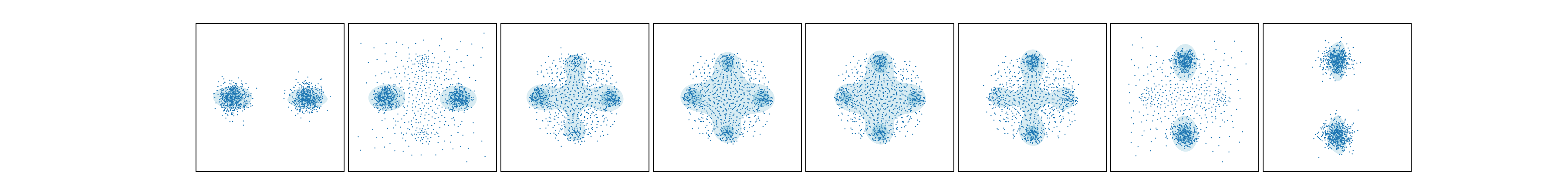} \\[0.3em]
			\includegraphics[width=\textwidth]{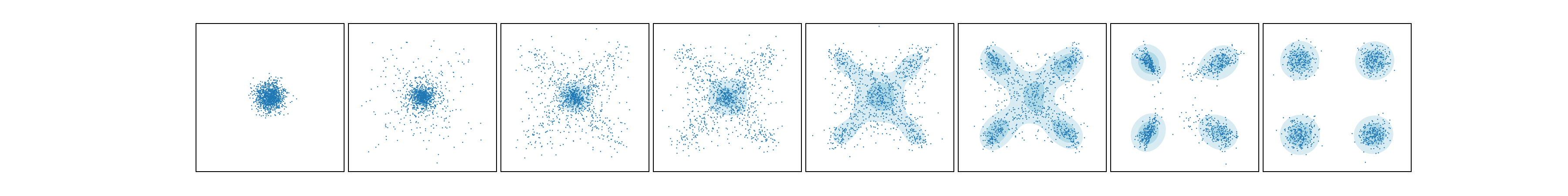} \\[-0.7em]
			\includegraphics[width=\textwidth]{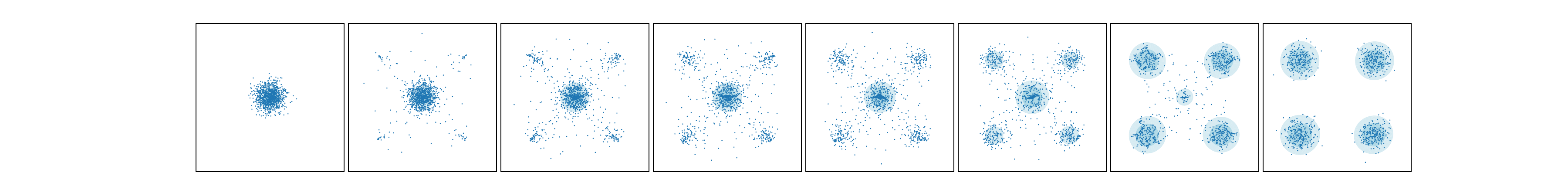} \\[-0.7em]
			\includegraphics[width=\textwidth]{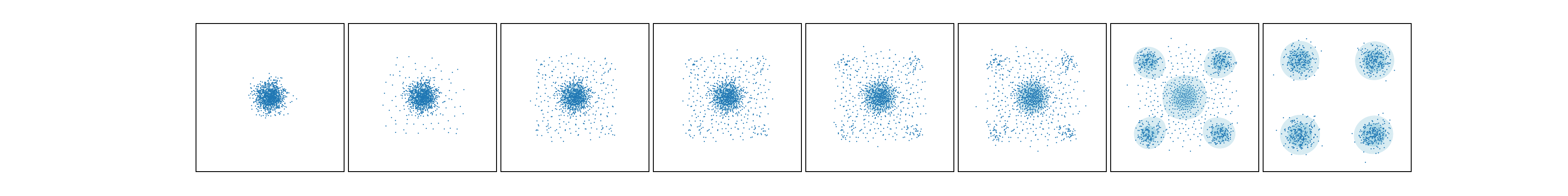}
		\end{center}
		\caption{Examples of interpolations with inverse multi-quadratic kernel \cref{eq.kernel_multiquadratic} with $c = 100, 10, 0.01$ (from top to bottom respectively). We show the last two examples from the experiment in \cref{fig:space_measure} in the main paper.}
		\label{fig:space_measure2}
	\end{minipage}
\end{figure}

	\Cref{fig:space_measure1} shows the last two examples from experiment shown in \cref{fig:space_measure} in the main paper, for which we construct interpolations using our model for different values of parameter $\gamma = 1/4, 1/16, 1/64$ in kernel \cref{eq.kernel_gamma}. Moreover, we consider the multi-quadratic kernel for different values of parameter $c = 0.01, 10, 100$. \Cref{fig:space_measure2} presents the results of interpolations with this kernel.
	
	This shows that the our approach results are irrespective of the choice of the actual kernel.

\section{Additional results from models Glow and Pioneer}\label{app.add_results}

\begin{figure}[t]
	\centering
	\includegraphics[width=0.95\textwidth]{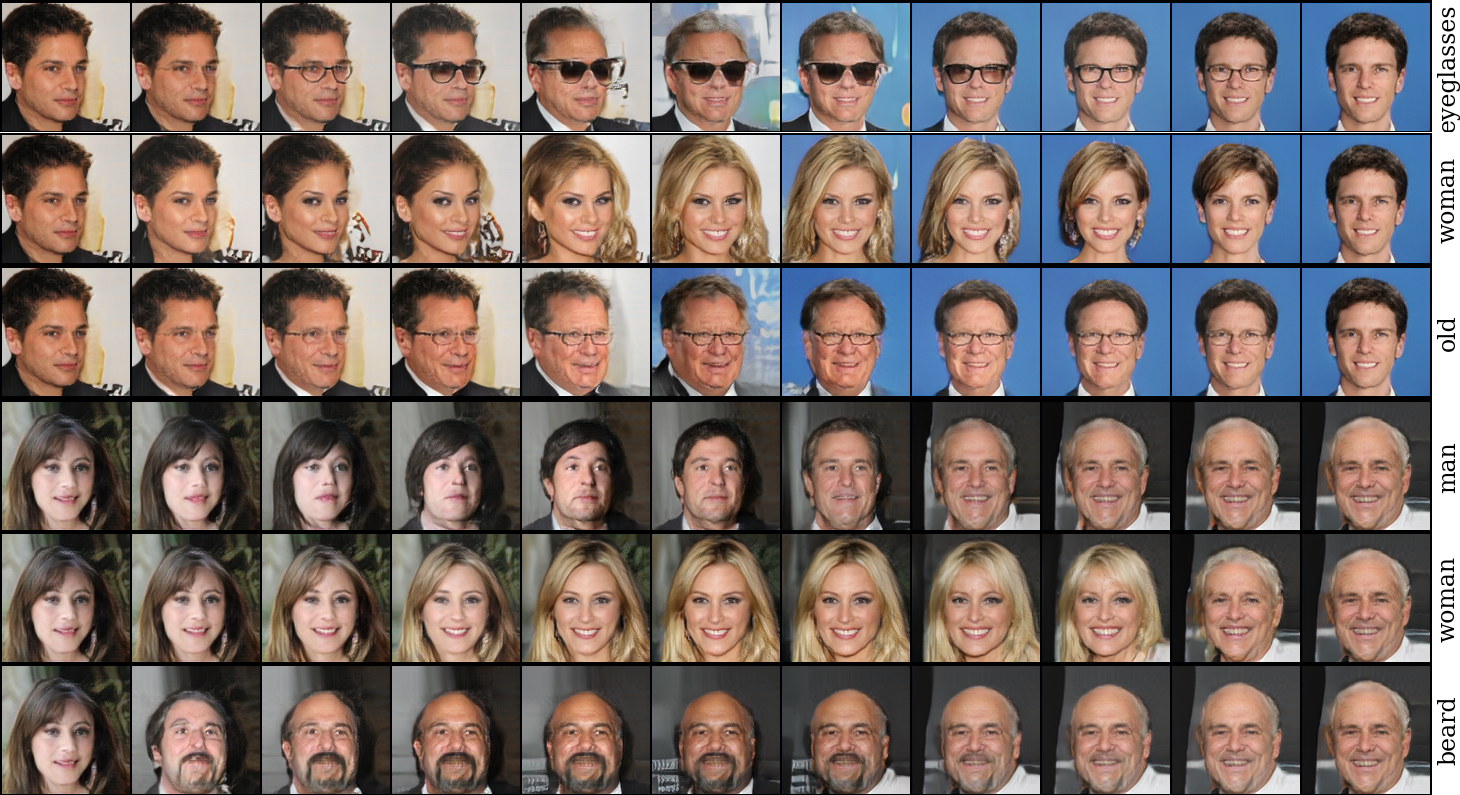}
	\caption{Proposed interpolation using a~pretrained Pioneer model between two different images trained to select images with few features. 
	}
	\label{fig:pioneer_more}
\end{figure}

\paragraph{Experiments with the pretrained Pioneer model}
In this section we present experiments were the endpoints selected for interpolation corresponded to different images, which corresponds to minimisation of energy in \cref{eq:cost} in the main paper. \Cref{fig:pioneer_more} shows the results.

\begin{figure}[h]
	\centering
	\renewcommand{\arraystretch}{0.1}
	\begin{tabular}{ @{}r@{}r@{} }
		\includegraphics[width=.94\textwidth]{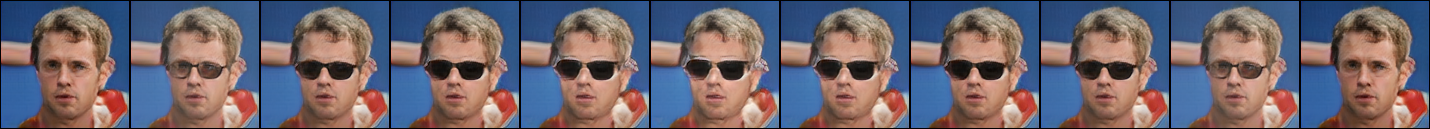}
		& \rotatebox{90}{\makebox[40pt][c]{\small eyeglasses}} \\
		\includegraphics[width=.94\textwidth]{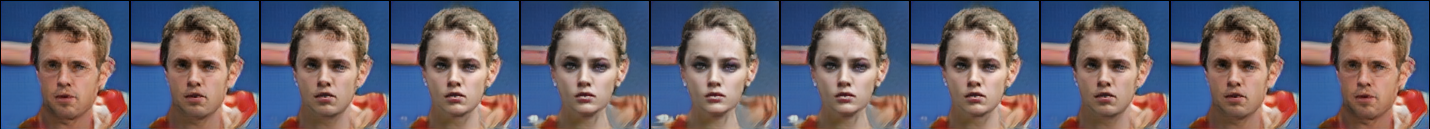}
		& \rotatebox{90}{\makebox[40pt][c]{\small young}} \\
		\includegraphics[width=.94\textwidth]{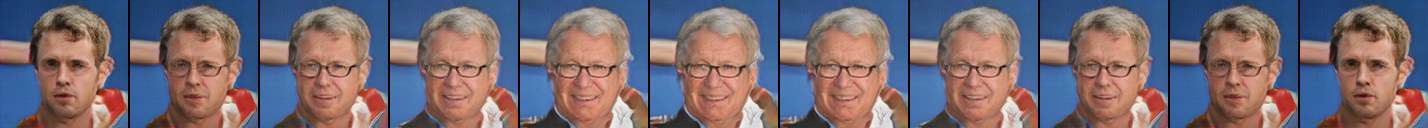}
		& \rotatebox{90}{\makebox[40pt][c]{\small old}} \\
		\includegraphics[width=.94\textwidth]{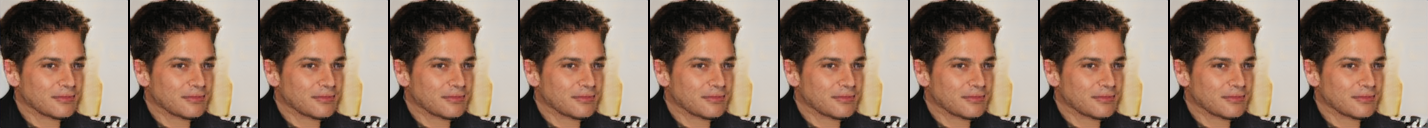}
		& \rotatebox{90}{\makebox[40pt][c]{\small man}} \\
		\includegraphics[width=.94\textwidth]{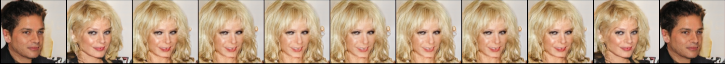}
		& \rotatebox{90}{\makebox[40pt][c]{\small blond hair}} \\
		\includegraphics[width=.94\textwidth]{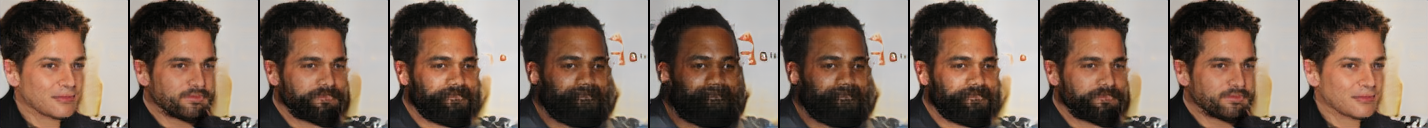}
		& \rotatebox{90}{\makebox[40pt][c]{\small beard}} \\
	\end{tabular}
	\caption{Examples of our interpolation on Pionier model between the same image from CelebA dataset, which were trained to find images with the selected features.
	}
	\label{fig:pionier_same_endpoints_more}
\end{figure}

\cref{fig:pionier_same_endpoints_more} presents some additional experiments where the defined endpoints were identical $z_0=z_1$, i.e. corresponded to the same image. In this case, an energy function with time element was used, see \cref{eq:cost_with_time} in the main paper. The addition of a $(t_{i+1}-t_i)$ term pushes the curve not to stay in the same place, even though that would be the shortest path. 

\begin{figure}[h]
	\centering
	\includegraphics[width=0.95\textwidth]{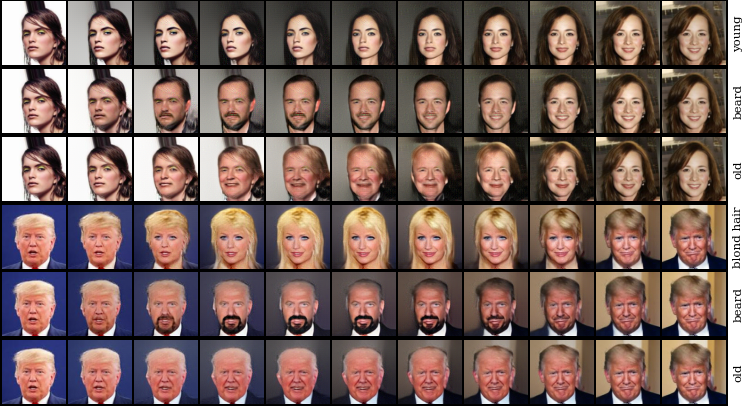}
	\caption{Our interpolation using a~Glow model between two President Donald Trump's images and other ones. The interpolations were to favour the beard, blond hair, old, young features. 
	}
	\label{fig:glow_diff_endpoints_more}
\end{figure}

\paragraph{Experiments with the pretrained Glow model}
\cref{fig:glow_diff_endpoints_more} presents examples for the Glow model interpolation with the proposed framework for the case where the endpoints are different. Again some attribute optimisations are susceptible to be modified with other attribute optimisations, especially the \emph{blond hair} frequently changes the gender. On the other hand, other attributes work perfectly, and the attributes are changed very well.

\begin{figure}[h]
	\centering
	\includegraphics[width=0.95\textwidth]{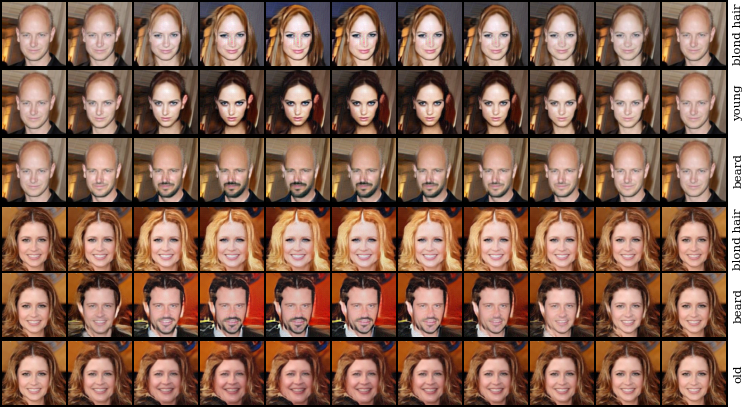}
	\caption{Examples of our interpolation on Glow model between the same image from CelebA dataset, which were trained to find images with the selected features.
	}
	\label{fig:glow_same_endpoints_more}
\end{figure}

\cref{fig:glow_same_endpoints_more} shows results for the Glow~\cite{Kingma2018glow} architecture with the same endpoints selected. The Glow generative model was in no way (just as in the case of the Pioneer) modified or optimised for the path selection.


\clearpage

\bibliographystyle{unsrt}  
\bibliography{refer}

\end{document}